\documentclass[preprint,3p,a4paper]{elsarticle}
\usepackage{setspace} 
\usepackage{amssymb}
\usepackage{amsmath}

\usepackage{subcaption}
\usepackage{hyperref}
\usepackage{siunitx, booktabs}
\usepackage{pgfplots}
\pgfplotsset{width=10cm,compat=1.9}
\usepackage{colortbl}
\usepackage{multirow}
\usepackage{booktabs}
\usepackage{pifont} 
\usepackage{rotating}
\usepackage{threeparttable} 
\usepackage{tablefootnote}
\journal{arXiv}

\begin{document}

\begin{frontmatter}



\title{Language Models for Business Optimisation with a Real World Case Study in Production Scheduling} 


\author[1]{Pivithuru Thejan Amarasinghe} 
\ead{p.amarasinghe@latrobe.edu.au}

\affiliation[1]{organization={Research Center for Data Analytics and Cognition},
            addressline={La Trobe University}, 
            city={Melbourne},
            postcode={3086}, 
            state={Victoria},
            country={Australia}}

\author[2]{Su Nguyen} 
\ead{su.nguyen@rmit.edu.au}

\affiliation[2]{organization={College of Business and Law},
            addressline={RMIT University}, 
            city={Melbourne},
            postcode={3000}, 
            state={Victoria},
            country={Australia}}

\author[1]{Yuan Sun} 
\ead{yuan.sun@latrobe.edu.au}

\author[1]{Damminda Alahakoon} 
\ead{d.alahakoon@latrobe.edu.au}

\begin{abstract}
Business optimisation has been used extensively to determine optimal solutions for challenging business operations. Problem formulation is an important part of business optimisation as it influences both the validity of solutions and the efficiency of the optimisation process. While different optimisation modelling languages have been developed, problem formulation is still not a trivial task and usually requires optimisation expertise and problem-domain knowledge. Recently, Large Language Models (LLMs) have demonstrated outstanding performance across different language-related tasks. Since problem formulation can be viewed as a translation task, there is a potential to leverage LLMs to automate problem formulation. However, developing an LLM for problem formulation is challenging, due to limited training data, and the complexity of real-world optimisation problems. Several prompt engineering methods have been proposed in the literature to automate problem formulation with LLMs. While the initial results are encouraging, the accuracy of formulations generated by these methods can still be significantly improved. In this paper, we present an LLM-based framework for automating problem formulation in business optimization. Our approach introduces a method for fine-tuning cost-efficient LLMs specifically tailored to specialized business optimization challenges. The experiment results demonstrate that our framework can generate accurate formulations for conventional and real-world business optimisation problems in production scheduling. Extensive analyses show the effectiveness and the convergence of the proposed fine-tuning method. The proposed method also shows very competitive performance when compared with the state-of-the-art prompt engineering methods in the literature when tested on general linear programming problems.
\end{abstract}




\end{frontmatter}



\section{Introduction}
\label{sec:introduction}

Business optimisation is an important process to help businesses gain competitive advantages by reducing operational costs, improving customer satisfaction, and mitigating risks. Advances in digital technologies, such as Internet-of-Things and cloud technologies, have enabled new business models with complex operations. Optimising key business decisions (operational, tactical, and strategic) in complex and dynamic systems is challenging and requires the involvement of different stakeholders. Handling business rules and various practical constraints is not a trivial task. Although modern optimisation technologies have offered businesses different ways to formulate and solve their problems, successfully adopting these technologies still requires significant domain knowledge and optimisation expertise \citep{drakeley2022inventory, ahmed2024lm4opt}.

Optimisation problems are various challenges from domains such as operations, economics, engineering, and computer science. Translating such a real-world challenge into a mathematical model \citep{antoniou2007practical} is usually done by optimisation experts. A constructed mathematical model can be solved by a solver. There are many efficient advanced solvers such as Gurobi \citep{gurobi}, Google OR-Tools \citep{cp-sat}, and CPLEX \citep{cplex2009v12}. However, a solver cannot automatically understand a mathematical model. Modelling languages help solvers in this process by providing language-specific syntax to describe a constructed mathematical model to a solver \citep{boyd2004convex}. The above process of constructing a problem formulation is time-consuming in operations research that requires expert knowledge. Inappropriate problem formulations can lead to infeasible solutions (e.g., failure to address constraints and optimise the objective of interest) and significantly slow down the solving process.

Large Language Models (LLMs) have gained popularity due to their broad range of applications. Originally introduced with the transformer architecture \citep{vaswani2017attention} for machine translation, LLMs have rapidly been adopted across various domains and business functions, including supply chain management \citep{li2024generative}, business analytics \citep{cheng2023gpt4}, marketing \citep{rivas2023marketing}, and programming \citep{nguyen2022empirical}. Companies such as Salesforce utilize code-generating LLMs to enhance developer productivity \citep{le2022coderl}. GitHub’s Copilot \citep{nguyen2022empirical} provides code suggestions and completions, while Amazon CodeWhisperer \citep{yeticstiren2023evaluating} assists developers in creating AWS resources using programming languages. Additionally, LLMs empower non-technical users to perform technical tasks, such as building simple websites or writing database queries, making advanced capabilities more accessible.

Constructing a problem formulation can be viewed as a language-to-language translation task, where a problem description in natural language is translated into a formulation expressed in a modelling language. This makes code-generating LLMs suitable candidates for fine-tuning in this domain. Early-stage code-generating LLMs faced limitations due to their reliance on labelled training data. However, the introduction of unlabelled data for training \citep{chen2021evaluating} has significantly mitigated these challenges. Modern pre-trained LLMs can now be fine-tuned for specific tasks with as few as a few hundred data points \citep{solaiman2021process}, a key characteristic that benefits this research. Despite these advancements, LLM-based applications in complex decision-making scenarios remain limited. Existing code-generating LLMs, primarily trained on generic programming problems, struggle with constructing problem formulations due to the need to represent intricate constraints, diverse optimization requirements, and various optimization techniques. Additionally, their scalability and resource demands pose significant challenges to their practical implementation.

Recently efforts have been made on automating problem formulation using LLMs. \citet{xiao2023chain} introduce a multi-agent-based prompt engineering technique with pre-trained LLMs to formulate complex operations research problems in supply chain management. \citet{lawless2023want} introduce a single agent-based prompt engineering technique with LLMs to formulate scheduling problems. These scheduling problems are about organising daily meetings of users. The users can provide their preferences in natural language to the system. The system creates daily schedules for the users using Constraint Programming (CP). In \citep{wasserkrug2024large} a research manifest is proposed to auto-formulate decision optimisation problems. However, these existing solutions have limitations such as relying on prompt engineering or in-context learning techniques, use of commercial language models, and scalability problems. Furthermore, commercial LLMs do not contain knowledge related to complex optimisation problems as they have been trained on generic data. Finally, LLMs have an input limit they can take and an output limit that they can generate. Therefore, solutions need to be scalable to construct complex optimisation problems.

We introduce a framework to automate problem formulation for complex real-world optimisation problems. The framework is based on finetuning pre-trained LLMs. We use an existing code-generating LLM as the pre-trained model. The idea behind using a pre-trained model is to minimise training data, training time, and training resources. We apply prompt engineering and modularisation techniques as our pre-trained model has a relatively small number of parameters. We measure the accuracy of the fine-tuned LLM by solving the constructed formulations using a solver and comparing final solutions. Therefore we ensure that the constructed formulations are error-free. To develop the training dataset, we introduce another framework in which users can create different problem descriptions using commercially available LLMs and manually construct their problem formulations. In contrast to existing solutions, our framework can generate formulations for complex optimization problems with limited resources while effectively addressing the scalability requirements of the formulation process. In the case studies, our framework constructs formulations of conventional and real-world production scheduling problems  \citep{xiong2022survey} using affordable computational resources. 

The contributions of this paper to automatically construct formulations of optimisation problems can be highlighted as follows:
\begin{itemize}
    \item Developing a framework for fine-tuning cost-efficient LLMs to automate problem formulation effectively.
    \item Developing a modularisation and prompt engineering technique to handle complex problem formulation.
    \item Developing two open-source datasets for conventional job shop scheduling problems and a real-world production scheduling scenario.
    \item Showing the efficacy of the proposed framework with conventional and real-world optimisation problems and benchmarking it against state-of-the-art prompt engineering approaches using an existing linear program dataset~\citep{ramamonjison2022augmenting}. 
\end{itemize}

The remainder of this paper is organised as follows. In Section \ref{literature}, we present past research in this research area. In Section \ref{cases}, we describe our two case studies. Sections \ref{methodology} and \ref{experiments} present our methodology and results. Finally, our conclusions are presented in Section \ref{conclusion}.

\section{Literature Review}
\label{literature}

In this section, we present a comprehensive analysis of past literature relevant to this research area. Starting with business optimisation and the process of solving optimisation problems, we move to the latest developments in LLMs. We present how LLMs have been leveraged in the optimisation domain and existing solutions in automating problem formulation. Finally, we present current challenges and opportunities for improvement in automating problem formulation.  

\subsection{Business Optimisation and Optimisation Technologies}

Business optimisation has been described as the Philosophy of Continuous Improvement \citep{singh2009kaizen}. In business optimisation, businesses attempt to make their operation as perfect as possible. This research considers business optimisation from a computational and mathematical perspective. In which businesses try to minimise or maximise an important characteristic of a process by an appropriate choice of decisions \citep{kallrath1997business}. Combinatorial optimisation is a class of optimisation used for mathematical and computational requirements of business optimisation \citep{yu2013industrial}.  Combinatorial optimisation is needed in real-world optimisation problems such as production scheduling \citep{pochet2006production}, vehicle routing \citep{toth2002vehicle},  bin-packing \citep{coffman1978application}, cut waste reduction \citep{klosowski2018integer}, and many more. To solve such problems algorithmic techniques such as dynamic programming, branch and bound, random-restart hill climbing, simulated annealing, genetic algorithms, and tabu search are developed. However, selecting a suitable algorithmic technique for a particular problem is a challenging task that requires human expertise.

\subsection{Problem Formulation and Solvers}

An optimization problem is formulated using an objective function, a set of constraints, and a set of variables. Optimization languages provide a structured syntax to represent mathematical models based on problem descriptions. Notable examples include MiniZinc \citep{nethercote2007minizinc}, GAMS \citep{soroudi2017power}, and AMPL \citep{fourer1990ampl}. Once formulated, these models are solved using optimization solvers, which vary in computational efficiency and solution strategies. Gurobi \citep{gurobi} is the state-of-the-art commercial solver, capable of handling a broad range of problems, including linear programming and mixed-integer programming. SCIP \citep{bestuzheva2021scip} is a leading non-commercial solver designed for mixed-integer programming and mixed-integer nonlinear programming. Modelling languages and solvers differ in their solving capabilities, licensing models, syntax expressiveness, and available documentation. Consequently, transforming a problem description from natural language into a structured formulation using a specific optimization language is a time-consuming process that requires significant human expertise.

\subsection{Large Language Models and Code Generation}

LLMs can perform a wide range of tasks such as reasoning \citep{peifeng2024joint}, writing academic literature \citep{lund2023chatgpt}, question answering \citep{wang2019multi}, language translation \citep{openai2023gpt4}, code generation \citep{le2022coderl}, and many more. LLMs use transformer-based architectures with self-attention \citep{vaswani2017attention} compared to predecessors like RNNs and LSTMs.  \citet{zan2022neural} report a comprehensive study on twenty-seven code-generating LLMs. Initially code-generating LLMs were trained using labelled datasets \citep{mastropaolo2021studying}. These LLMs had limitations as such techniques had practicality issues with preparing a dataset. With the introduction of unlabelled data to train LLMs for code generation by \citet{chen2021evaluating} limitations have been reduced. \citet{chen2021evaluating} is a code-generating LLM fine-tuned using a large corpus of GitHub Python code. Therefore it can cover a broader spectrum. \citet{le2022coderl} have improved the quality of generated code by considering the results of test execution with reinforcement learning. \citet{zhang2023planning} introduce a model-agnostic planning process based on Markov's decision process.

Benchmarks allow us to find progressive improvements in the code-generating capabilities of LLMs. APPS \citep{hendrycks2021measuring} benchmark includes $10,000$ programming problems and their solutions in Python. It has simple and complex problem descriptions closer to the natural language. HumanEval \citep{chen2021evaluating} benchmark includes $164$ hand-written programming problems to measure the functional correctness of generated code. Compared to APPS, HumanEval does not contain solutions in GitHub. MBPP \citep{austin2021program} benchmark includes $974$ Python programming problems, which suit entry-level programmers. Metrics like $pass@k$ \citep{kulal2019spoc} have been introduced to evaluate the accuracy of the generated code. For a particular problem description, $k$ number of solution codes are generated. Generated codes are run against a test case related to a programming problem and the programming problem is considered solved if at least one solution code out of $k$ can pass the test case.

\subsection{Large Language Models for Optimisation}

Recently, efforts have been made to leverage LLMs in the optimisation domain for algorithmic improvements and to automate the construction of problem formulations. Considering algorithmic improvements, \citet{yang2023large} introduce an LLM-based approach to solve the absence of gradient in derivative-based algorithms. \citet{liu2023large} study about how LLMs can reduce expertise needed when designing operators for evolutionary algorithms. Focusing on genetic programming \citet{lehman2022evolution} utilise LLMs trained on code generation to improve the effectiveness of mutation operators.  In \citep{bradley2024openelm} a new open-source algorithm is introduced to design evolutionary algorithms, which uses LLMs in operations such as variation, fitness evaluation, and diversity measurements.

Early research on automating problem formulations initiated from the NL4Opt dataset \citep{ramamonjison2023nl4opt}. Researchers have used the NL4Opt dataset to fine-tune LLMs to construct problem formulations and as a benchmark to evaluate their solutions. In the NL4Opt dataset, there are problem descriptions and linear programming problem formulations related to them. However, these problem descriptions are simpler compared to real-world optimisation problems and respective problem formulations contain on average $2.08$ variables and $2.83$ constraints per problem formulation. Based on the NL4Opt dataset, \citet{abdullin2024synthetic} introduce a multi-agent-based approach for linear programming problems. As agents, researchers use LLMs with prompt engineering techniques. One agent makes conversations with the user to extract important information related to the problem. Based on the information, another agent constructs the related problem formulation. \citet{li2023synthesizing} fine-tune LLMs to formulate Mixed-Integer Linear Programming (MILP) models for unstructured natural language descriptions of decision problems. The dataset used for fine-tuning comprised thirty problems with their MILP models. \citet{lawless2023want} combine LLMs with Constraint Programming (CP) to facilitate interactive decision support. They use pre-trained LLMs and prompt engineering techniques to interact with LLMs. To automate the construction of problem formulations for complex operations research (OR) problems, \citet{xiao2023chain} introduce a multi-agent-based approach by leveraging LLMs. Named as Chain-of-Experts (CoE), this framework has agents performing specific roles. Each agent has OR knowledge related to the tasks they perform. As the dataset, this framework has introduced a new dataset with more complex OR problems compared to the NL4Opt dataset. The proposed framework uses prompt engineering and in-context learning techniques to automate the construction of problem formulations from problem descriptions. \citet{ahmaditeshnizi2023optimus} introduce OptiMUS, an LLM-based agent designed to formulate and solve MILP problems from their natural language descriptions. \citet{wasserkrug2024large} propose to develop a Decision Optimisation CoPilot (DOCP) by leveraging LLMs and optimisation to assist decision-makers. The users can interact with the proposed system in natural language and the system can grasp the business problem, subsequently formulating and solving the corresponding optimisation model. For the proposed system researchers identify three key requirements. They are the ability to translate a business-level problem definition to an optimisation model, the ability to verify the correctness of the model and the ability to create efficient models.

A comprehensive analysis of existing LLM-based problem formulation solutions is presented in Table \ref{table:existing_application_comparison}. This analysis examines three key aspects of these solutions: the automated optimization model, the methodology employed with LLMs, and the evaluation metric used. Regarding optimization models, existing solutions cover Linear Programming (LP), Constraint Programming (CP), and Mixed Integer Programming (MIP). In terms of LLM methodology, they leverage three primary techniques commonly used in NLP tasks with LLMs: prompt engineering, in-context learning, and fine-tuning. For evaluation, these solutions assess either the correctness of the mathematical formulation or the final output of the problem.

\begin{sidewaystable}[!htbp]
\centering
\captionsetup{justification=centering}
\caption{Existing LLM-Based Applications for Automating Problem Formulation}
\small
\renewcommand{\arraystretch}{1.2}
\begin{tabular}
{>{\centering\arraybackslash}p{0.18\linewidth}  >{\centering\arraybackslash}p{0.03\linewidth} >
{\centering\arraybackslash}p{0.04\linewidth} >
{\centering\arraybackslash}p{0.05\linewidth} >
{\centering\arraybackslash}p{0.08\linewidth} >
{\centering\arraybackslash}p{0.08\linewidth} >
{\centering\arraybackslash}p{0.05\linewidth} >
{\centering\arraybackslash}p{0.04\linewidth} >
{\centering\arraybackslash}p{0.04\linewidth} >
{\centering\arraybackslash}p{0.09\linewidth} >
{\centering\arraybackslash}p{0.09\linewidth}}
 \toprule
\multirow{6}{*}{Paper} 
& \multicolumn{3}{c}{\shortstack{Optimisation\\Model}} 
& \multicolumn{5}{c}{LLM Methodology}
& \multicolumn{2}{c}{\shortstack{Evaluation\\Metric}}\\
\cmidrule[1.0pt](lr){2-4} \cmidrule[1.0pt](lr){5-9} \cmidrule[1.0pt](lr){10-11}

& LP & CP & MIP 
        & Prompt Engineering & In-Context Learning & Fine Tuning & \multicolumn{2}{c}{\shortstack{Agents/ \\Workflows}} & \shortstack{Semantic\\Based Metric\tablefootnote{The objective function, constraints, and variables are evaluated against ground truth data.}} & \shortstack{Pass Rate\\Based Metric}\\
\cmidrule[1.0pt](lr){8-9}        
& & & & & & & Single & Multi  \\
 \toprule
\citet{xiao2023chain} & \checkmark & & \checkmark & \checkmark & \checkmark & & & \checkmark & & \checkmark\\

\citet{lawless2024want} & & \checkmark & & \checkmark & \checkmark & & \checkmark & & & \checkmark\\

\citet{li2023synthesizing} &  & & \checkmark & \checkmark & & & \checkmark & &  \checkmark & \checkmark\\

\citet{ahmaditeshnizi2023optimus} & \checkmark & & \checkmark & \checkmark & \checkmark & & \checkmark & & & \checkmark\\

\citet{tang2024orlm} & \checkmark & & \checkmark &  & & \checkmark & & & \checkmark & \checkmark \\

\citet{ahmed2024lm4opt} & \checkmark & & &  &  & \checkmark & & & \checkmark & \\

\citet{zhang2024solving} & \checkmark & & \checkmark & \checkmark & & \checkmark &  & \checkmark & \checkmark &  \\

\citet{mostajabdaveh2024optimization} & \checkmark & & \checkmark & \checkmark & & & & \checkmark & \checkmark &\\

\citet{wang2023opd} & \checkmark & & & & & \checkmark & &  & \checkmark &\\


\citet{wang2024leveraging} & & & \checkmark & \checkmark &  & \checkmark & \checkmark &  & & \checkmark\\

\citet{jiang2024llmopt} & \checkmark & & \checkmark & \checkmark & & \checkmark & \checkmark &  & & \checkmark\\

\citet{astorga2024autoformulation} & \checkmark & & \checkmark & \checkmark & & &  \checkmark & & \checkmark & \checkmark\\

\citet{deng24cafa} & \checkmark & & & \checkmark & \checkmark & & \checkmark &  & & \checkmark \\

\citet{dakle2023ner4opt} & \checkmark & & & \checkmark & \checkmark & \checkmark & &  & & \checkmark\\

\citet{wang2024bpp} & \checkmark  & & \checkmark & \checkmark & & \checkmark & & \checkmark & & \checkmark\\

\citet{mongaillard2024large} & \checkmark & & \checkmark & \checkmark & \checkmark & & & \checkmark & & \checkmark\\

\citet{ghiani2024integrating} & & & \checkmark & \checkmark & & & & \checkmark & & \checkmark\\

\citet{zhang2025decision} &  & & \checkmark & \checkmark & \checkmark & & & \checkmark & & \checkmark\\

\citet{li2025abstract} & \checkmark & & \checkmark & \checkmark & \checkmark & &  & \checkmark  & \checkmark & \checkmark\\


\citet{mahmud2025distributed} &  & \checkmark & & \checkmark & \checkmark & & & \checkmark & & \checkmark\\

\citet{jiao2025integrating} &  & & \checkmark & \checkmark & \checkmark & & \checkmark &  & \checkmark & \checkmark\\



 \bottomrule
\end{tabular}
\label{table:existing_application_comparison}
\end{sidewaystable}

\subsection{Challenges of Automating Problem Formulation with LLMs}

LLMs have inherent limitations such as the need for vast amounts of data and computational resources, high costs of training and maintenance, limited scalability, and lack of causality \citep{hadi2023survey}. The scalability aspect of LLMs is a major challenge when leveraging LLMs to optimisation problems. As scalability what we expect is the ability to expand an LLM-based solution developed to construct small problem formulations, to construct large and complex problem formulations. Training datasets, training time, training resources, and token (a word, a sub-word or a character) limits of the LLM contribute towards the scalability of the LLM. These factors have trade-offs between each other such as an increase in token limit can cause lengthier training time and more computational resources. Lack of causality in LLMs causes the need for larger datasets to represent complexities in optimisation problems. As most of the public datasets do not contain data related to optimisation problems, representing different optimisation techniques for complex optimisation problems will be challenging. In our framework, we use fine-tuning to provide optimisation knowledge to the pre-trained LLM. Fine-tuning is an effective technique compared to in-context learning / prompt engineering and a cost-efficient technique compared to training an LLM from scratch. To address the scalability, we use prompt engineering and modularisation techniques with cost-efficient pre-trained LLMs.

\section{Case Study} 
\label{cases}

In this section, we introduce the two case studies for our research, with Section~\ref{js} the case study of conventional job shop scheduling and Section~\ref{fjs} the case study of real-world production scheduling.

\subsection{Conventional Job Shop Scheduling}
\label{js}

As conventional production scheduling problems, we select Job Shop Scheduling (JSS) problems mentioned in the literature. JSS is one class of combinatorial optimisation problems that is common in manufacturing \cite{pinedo2005planning}. JSS has its variants such as Flexible Job Shop Scheduling (FJSS) \citep{fattahi2007mathematical} and Dynamic Job Shop Scheduling (DJSS) \citep{liu2024dynamic}. Due to its practical constraints and complex nature, JSS is one of the most popular NP-hard optimisation problems investigated by researchers in operational research and computer science. The goal of JSS is to schedule a number of jobs over a number of machines, and each job consists of a set of operations that need to be executed in a given order on the allocated machine. The machine is allowed to process one operation at a time, and different objectives such as makespan and weighted tardiness can be minimised.  JSS is solved using methods such as integer programming \citep{ku2016mixed}, metaheuristics \citep{kreipl2000large}, and constraint programming \citep{beck2011combining, watson2008hybrid}. The mathematical formulation related to JSS can be found in \ref{appendix:JSS}.

\subsection{Real-World Production Scheduling}
\label{fjs}

\begin{figure}[!htbp]
    \centering
    \includegraphics[width=0.95\linewidth]{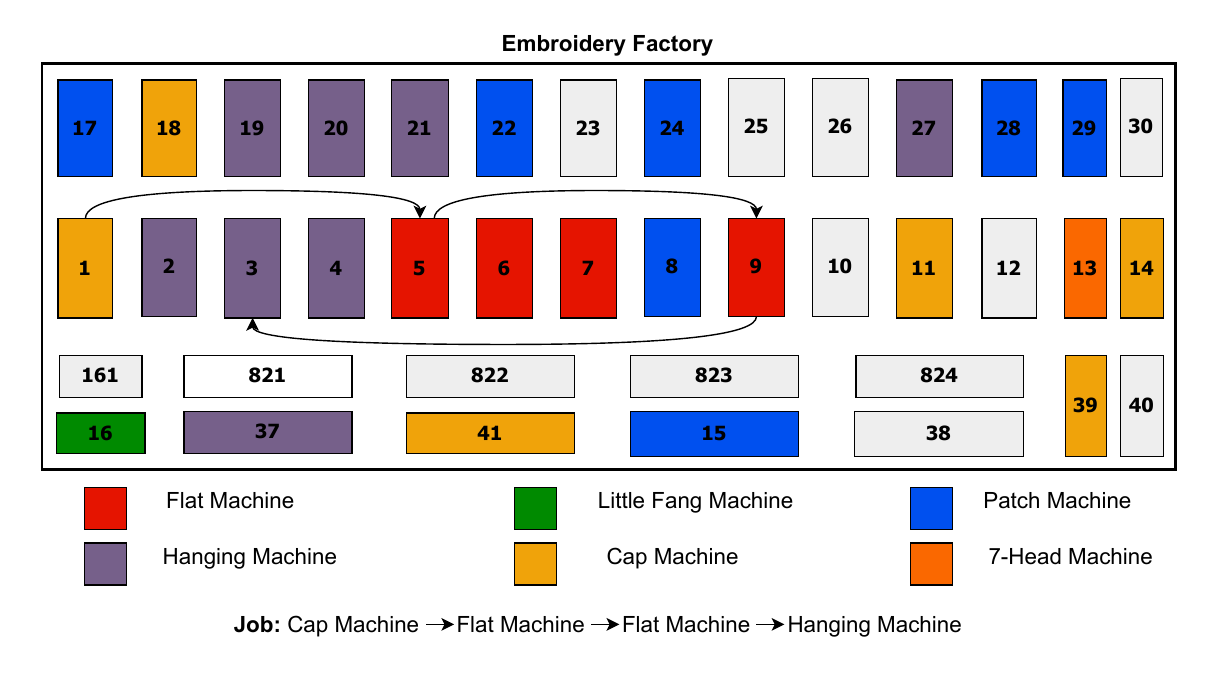}
    \caption{Layout of a Melbourne embroidery factory. In the diagram, the arrows indicate the sequence of machines used to process an actual job.}
    \label{fig:factory_layout}
\end{figure}

As real-world production scheduling problems, we tackle production scheduling problems in an embroidery producer in Australia, which we name company ``XYZ" in this paper. The clothing company has different printing and embroidery facilities across Australia. The embroidery factory in this case study has six machine groups (Figure \ref{fig:factory_layout}). Each machine group has one or more machines. The machine groups are Flat Machine, Hanging Machine, Cap Front, Patch Machine, Little Fang Machine, and 7-Head Machine. In addition to scheduling the jobs, the clothing company wants to schedule production machines. This makes the optimisation problem a Flexible Job Shop Scheduling (FJSS) problem. The mathematical formulation related to FJSS can be found in \ref{appendix:FJSS}

In the embroidery factory, a job consists of multiple operations. Each job upon entering the factory waits in the waiting area for the assignment. Jobs with less than four pieces are processed using the Little Fang Machine. The jobs with four to seven items are processed using the 7-Head Machine. If the waiting time for the 7-Head Machine exceeds two working days, the job is assigned to another machine. The current scheduling happens by sorting jobs based on the due date. However, the company's main Key Performance Indicator (KPI) is machine utilisation. Minimising makespan helps to improve machine utilisation. Besides minimising makespan, total tardiness is another objective that we need to optimise.

\section{Proposed Method}
\label{methodology}

\subsection{Overview}
 
\begin{figure}[!htbp]
    \centering
    \includegraphics[width=0.75\linewidth]{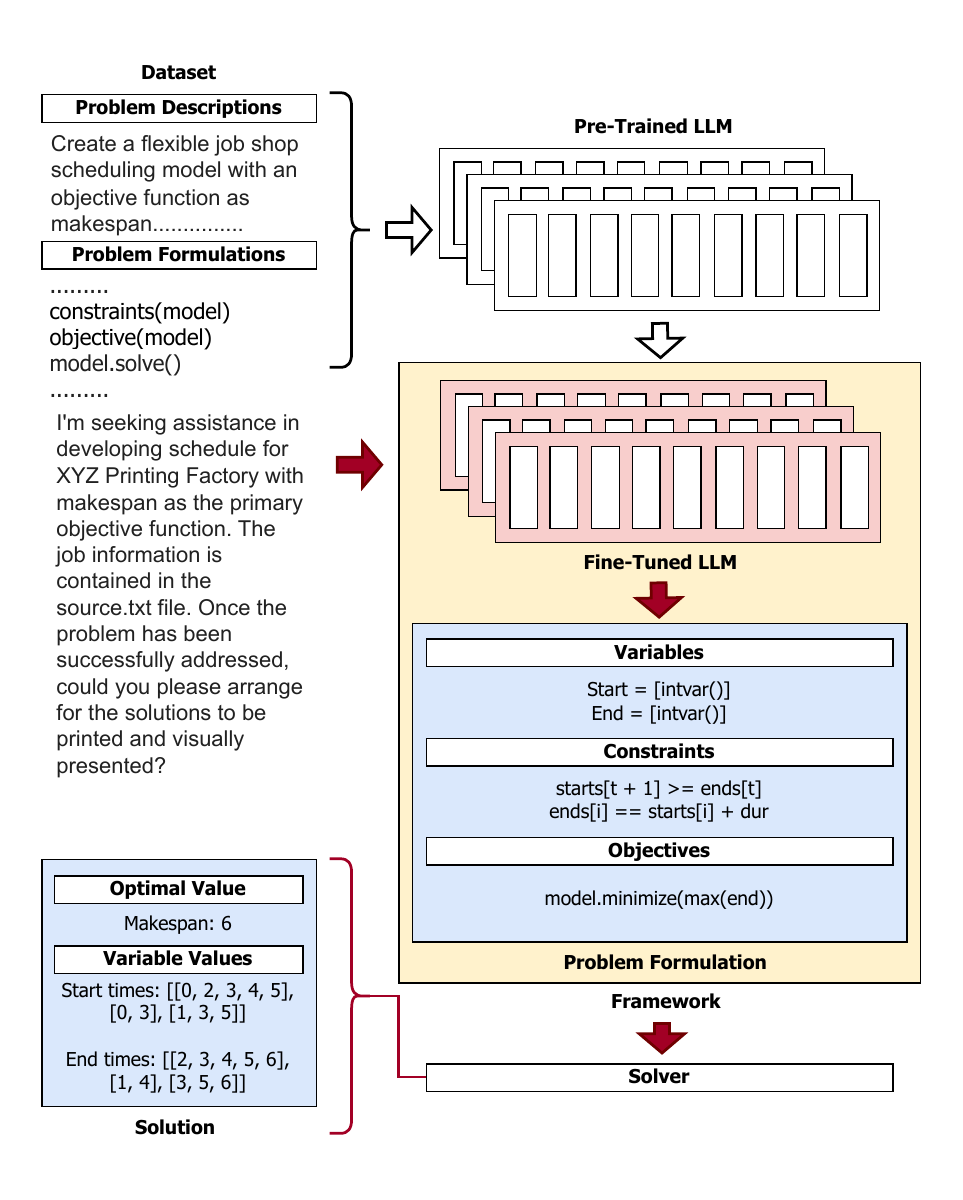}
    \caption{Overview of the proposed framework. In the framework, a pre-trained LLM is fine-tuned using problem descriptions and relevant problem formulations. The fine-tuned LLM is capable of constructing a problem formulation for a given problem description.}
    \label{fig:solution_overview}
\end{figure}

The remainder of this section provides a detailed overview of our framework. As illustrated in Figure \ref{fig:solution_overview}, the process begins with a dataset containing optimization problems and their corresponding formulations. This dataset is then used to fine-tune a pre-trained LLM, enabling it to generate problem formulations for given optimization problems.

\subsection{Dataset Development}

\begin{figure}[!htbp]
    \centering
    \includegraphics[width=0.75\linewidth]{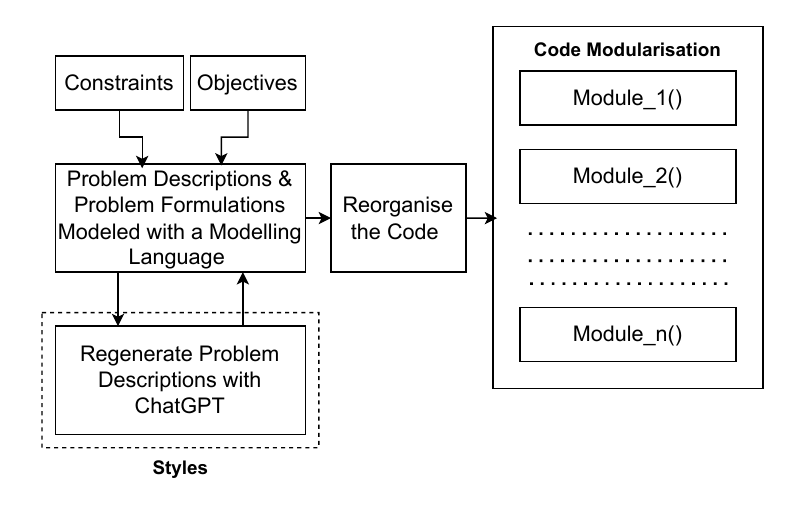}
    \caption{An overview of dataset development. Problem descriptions and problem formulations are created based on objectives and constraints. The problem descriptions are regenerated using ChatGPT in different styles. Finally, problem formulations are reorganised into code modules.}
    \label{fig:dataset_development}
\end{figure}

The first step is identifying the objective and constraints of the optimisation problem. Then problem descriptions need to be created. For the problem descriptions, respective problem formulations should be constructed using an optimisation language. In our experiments, we generated problem formulations for a constraint programming solver. However, the same methodology can be applied to other solvers such as mixed integer programming solvers. For the case study with conventional job shop scheduling problems, we conducted the dataset preparation process manually. However, manually doing this process is time-consuming. Therefore, in the real-world production scheduling case study, different problem descriptions were automatically generated (Figure \ref{fig:dataset_development}). ChatGPT \citep{openai2023gpt4} is utilised to modify existing problem descriptions with various styles \citep{reif2021recipe}. So different stakeholders can be represented. To fine-tune an LLM, a significantly small amount of data is needed compared to training. This can vary depending on the size of the LLM and the complexity of the problems. Once a sufficient number of data points are created, constructed formulations need to be reorganised into modules. This is to fulfil scalability requirements when constructing large problem formulations. In the scalability section, we will dive deeper into it.

\subsection{Fine-Tuning}

In fine-tuning, based on the problem descriptions in the dataset, the pre-trained model generates problem formulations. These generated problem formulations are compared with target problem formulations. Based on the comparisons an error is calculated. The weights of the pre-trained model are adjusted to reduce this error. This adjustment happens via the optimisation algorithm used by the pre-trained model. The architecture of the pre-trained model determines the error calculation mechanism and optimisation algorithm. Until we get a satisfactory error this process is repeated for a few iterations. The number of iterations needed in fine-tuning a pre-trained model is relatively low compared to training a model from scratch. Therefore, using a pre-trained model reduces training time and computational resources. A suitable pre-trained model should have excellent code-generation skills because optimisation languages have similar syntax and behaviour to programming languages. Similarly to code-generation LLMs that construct programming problems, a fine-tuned code-generation LLM can construct problem formulations. For the experiments in the framework, CodeRL \citep{le2022coderl} is used as the pre-trained model. CodeRL is an open-source model that has a comparatively smaller parameter size. However, it has effective code-generation capabilities, due to its reinforcement learning-based training. CodeRL is an extension of CodeT5 \citep{wang2021codet5} and has an encoder-decoder transformer model with self-attention. As training data, it uses GitHub data.

Although CodeRL is showing promising results with code generation, it faces challenges in constructing problem formulations because CodeRL's training dataset does not contain data related to problem formulations. Even GPT-4 \citep{openai2023gpt4} finds it difficult to construct complex formulations as its training data does not contain relevant examples.  Techniques such as in-context learning can be used to provide some knowledge on the construction of formulations for commercial LLMs, but that needs some optimisation expertise. Therefore, we fine-tune the pre-trained model, CodeRL, using a domain-specific dataset to improve its capability for constructing problem formulations. Another challenge faced by CodeRL is its token limit. CodeRL can generate an output with a maximum of 600 tokens. However, the production scheduling problems we experiment with the framework cannot be constructed using 600 tokens. In the next section, we address this challenge of managing token limits.

\subsection{Scalability}

LLMs can process a specific number of tokens at a time. This is known as the token limit of an LLM. Token limit defines the maximum number of input characters LLM can take and the maximum number of output characters LLM can generate. LLMs with higher token limits can generate larger outputs. However, a higher token limit means a larger parameter size for the LLM. A larger parameter size makes an LLM larger, which has a higher training and operational cost. Therefore, token limit and LLM size have a trade-off between each other. In the framework, we use cost-efficient smaller models. Cost-efficient models improve the practical use of our framework. However, a constructed problem formulation for an optimisation problem exceeds the token limits of a cost-efficient LLM. To address this, we use prompt engineering and modularisation.

While creating the dataset, we reorganised constructed problem formulations into modules. Such a module is self-sufficient, which means it has all the variables and methods to perform the desired task. Also, these code modules fit into the token limits of the LLM. To generate each code module, we provide an instruction (prompt) to the LLM in addition to the optimisation problem. The instructions can vary from optimisation problem to problem. Before fine-tuning the LLM, we can create our own instruction set based on the optimisation problem. The instructions we used when constructing our case studies can be seen in the experiment section. As a prompt, we append the desired instruction to the problem description of the optimisation problem. This makes the original optimisation problem into a set of sub-problems, each of which has its own code module. So the original dataset gets expanded by several instructions, and this new dataset is used to fine-tune the LLM. The fine-tuned LLM constructs a problem formulation instruction by instruction. In the end, once we combine instruction-wise code modules, we get the final formulation.

\subsection{Performance Metrics}

Traditionally, LLMs use cross-entropy-based loss calculation (Equations \ref{cross_entropy_loss_1}, \ref{cross_entropy_loss_2}). The cross-entropy loss \citep{mao2023crossentropylossfunctionstheoretical} is a reliable measurement used by current learning algorithms for classification tasks. Based on cross-entropy loss an LLM adjusts its weights during back propagation. In cross-entropy loss, LLMs compare a generated token with the desired token. By reducing this loss, LLMs learn to accurately predict the next token based on the past tokens.

\begin{equation}\label{cross_entropy_loss_1}
    l(x,y)=\sum\limits_{n=1}^Nl_n/N,
\end{equation}
\begin{equation}\label{cross_entropy_loss_2}
    l_n=-\log\frac{\exp(x_{n,y_n})}{\sum\limits_{q=1}^Q\exp(x_{n,q})}.1\{y_n\neq \text{ignore\_index}\} ,
\end{equation}
where $x$: logits from the LLM for a given problem description (generated problem formulation), $y$: target ids (target problem formulation), ignore$\_$index: target value that is ignored, $Q$: number of classes, $N$: mini-batch dimension.

In our framework, we use the same cross-entropy loss. During the fine-tuning process, LLM follows a cross-entropy loss-based learning process. Fine-tuned LLM learns optimisation knowledge that needs to be used for a particular optimisation problem based on cross-entropy loss. Different optimisation techniques learned by the LLM depend on the variety in the training dataset. However cross-entropy loss alone is insufficient to ensure the problem formulation constructing capability of an LLM. Therefore, we solve constructed problem formulations using a solver. For the experiments, we use OR Tools \citep{ortools} as the solver. Once a constructed problem formulation is solved using the solver, there can be three outcomes. Problem formulation can give the correct solution (success), an incorrect solution (failure), or an error during the execution (exception). This we call execution status. In addition to the cross-entropy loss, the framework uses execution status to evaluate the fine-tuning process of the LLM. The success, failure, and exception rates are defined as follows:

\begin{equation}\label{success_rate}
    Success = \frac{CR}{I} \times 100\%,
\end{equation}

\begin{equation}\label{failure_rate}
    Failure = \frac{RE}{I} \times 100\%,
\end{equation}

\begin{equation}\label{exception_rate}
    Exception = \frac{CE}{I} \times 100\%,
\end{equation}
where $I$: total number of problem formulations, $CR$: number of problem formulations that gave the correct solution, $RE$: number of problem formulations that gave the incorrect solution, $CE$: number of problem formulations that fail to run. The execution status evaluates the correctness of the constructed problem formulations.

\section{Experiments}
\label{experiments}

We conduct experiments on our framework with conventional and real-world production scheduling problems. Later we expand our framework to a linear programming dataset LPWP \citep{ramamonjison2022augmenting} to compare its performance with existing methods. Once the datasets are created, we randomly allocate $70\%$ of each dataset as training data, $10\%$ as validation data, and $20\%$ as test data. The following sections will provide additional information on our experiments.

\subsection{Conventional Production Scheduling Problems}

We manually created a hundred production scheduling problem descriptions with their problem formulations (Figure \ref{fig:dataset_development}), and the dataset has been made available on Github\footnote{\url{https://github.com/pivithuruthejanamarasinghe/AI-Copilot-Data}.}. We constructed problem formulations using the CPMpy \citep{guns2019increasing} library. 
The problem descriptions in the dataset have less than six hundred tokens. The respective problem formulations have tokens in the range of $1200$ to $1800$. A sample problem description can be,

\begin{quote}
``
Job shop scheduling model with $5$ jobs and $5$ machines. All jobs have random routes and their operations have random durations. The objective function is makespan. Maximum duration is $20$. After solving the problem, solutions will be printed and visualised. Note: The second task of Job two has to come before all the second tasks of other jobs.
"
\end{quote}

To fulfil the scalability requirements,  we reorganised constructed problem formulations to code modules. As instructions, we use nine prompts. Correspondingly each problem description becomes nine different problem descriptions. Therefore, our dataset consists of 900 instances. To fine-tune the pre-trained model, we use a trainer \citep{huggingface-trainer}. Fine-tuning configurations are available in Table \ref{table:traing_configurations}. We pick batch size and epoch count as primary parameters for hyperparameter tuning. The batch size is the number of data points we consider for each batch. The epoch count is the number of training iterations that we conduct. The tuning of these two parameters has already generated promising results. Therefore, we did not tune other parameters such as learning rate.

\begin{table}[!htbp]
\centering
\captionsetup{justification=centering}
\caption{Training Configurations}
\begin{tabular}
{p{0.35\linewidth}  p{0.5\linewidth}} 
\toprule
 \textbf{Training Configuration} & \textbf{Value} \\ [0.5ex] 
\midrule
GPU type & NVIDIA Tesla V100 SXM2 $32$ GB\\
Pre-trained model & Salesforce/codet5-large-ntp-py\\
Tokenizer & Salesforce/codet5-large-ntp-py\\
Learning rate & $5e-05$\\
Gradient checkpointing & True\\
Evaluation strategy & Steps\\
Evaluation steps & $10$\\
Logging steps & $10$\\
Do Evaluation & True\\
\bottomrule
\end{tabular}
\label{table:traing_configurations}
\end{table}

As evaluation metrics, we use cross-entropy loss, training time, success rate, failure rate, and exception rate. Loss values are low for all batch sizes (Table \ref{table:train}). The batch size has a significant impact towards training time as batch size controls the frequency that we update the weights of the LLM. After each batch, the fine-tuning process adjusts the weights of the LLM to reduce the loss. For smaller batch sizes, training time is high as there are more frequent weight updates. With respect to success rates, all the batch sizes have been able to achieve more than 95\% success rate. Having good success rates for higher batch sizes is cost-beneficial as it reduces the training time and computational cost. With the increase in the epoch count, we observe that loss is reduced. However, an increase in the epoch count adversely affects training time. The training time increases with the epoch count because the epoch count determines the number of training iterations. With epoch count, success rate increases and failure and exception rates decline. Therefore, we can observe that with more training, the pre-trained LLM can perform better in constructing problem formulations. We observe that batch size two and epoch count four or batch size four and epoch count four is a more desirable training setting as they can produce good training results with lower training time.

Using the validation dataset we investigate the loss convergence during the fine-tuning process and check for over-fitting. For all parameter settings, validation and training loss curves overlap at some point (Figure \ref{appendix:cs1}). That indicates our fine-tuning process has not been an overfitting scenario. For batch size one, training and validation lost curves are similar. In batch size two, for initial epoch counts there is a small difference between training and validation loss curves. However, that difference reduces with the increase of the epoch count. For batch size four, the difference between the training and validation curve is significant for initial epoch counts. However, this difference reduces for higher epoch counts similar to the other batch sizes. Settings with batch size two and epoch count four or batch size four and epoch count four are more suitable settings for fine-tuning because they have small differences between curves and the curves are not identical. 

Using testing data we measure success rate, failure rate, and exception rate (Figure \ref{fig:test}). In all batch sizes, we observe success rates greater than 95\%. With the increase in epoch count, we observe a significant improvement in success rates. Except for batch size one, we obtained more than 95\% success rate after epoch count two. However, for batch size one we obtain a more than 95\% success rate at epoch count eight. The exception rate is high for lower epoch counts. Except in batch size one, the exception rate is lower than 5\% for epoch counts greater than two. However, for batch size one, we obtain an exception rate lower than 5\% at epoch count eight. 

\begin{table}[!htbp]
\centering
\captionsetup{justification=centering}
\caption{Training Results - Conventional Job Shop Scheduling}
\begin{tabular}
{>{\centering\arraybackslash}p{0.11\linewidth}  >{\centering\arraybackslash}p{0.06\linewidth} >
{\centering\arraybackslash}p{0.11\linewidth}  >{\raggedleft\arraybackslash}p{0.11\linewidth} >
{\raggedleft\arraybackslash}p{0.11\linewidth}  >{\raggedleft\arraybackslash}p{0.11\linewidth} >
{\raggedleft\arraybackslash}p{0.11\linewidth}}
 \toprule
{\textbf{Batch size}} & {\textbf{Epoch}} & {\textbf{Loss}} & {\textbf{Time(sec)}} & {\textbf{Success}} & {\textbf{Failure}} & {\textbf{Exception}} \\
 \midrule
 1 & 1 & 0.0056 & 1083.78 & 16\% & 23\% & 61\%\\
 1 & 2 & 0.0068 & 2105.51 & 0\% & 0\% & 100\%\\
 1 & 4 & 0.0038 & 4216.54 & 0\% & 1\% & 99\%\\
 1 & 8 & 0.0008 & 8561.05 & 96\% & 0\% & 4\%\\
\hline
 2 & 1 & 0.0363 & 675.19 & 0\% & 0\% & 100\%\\
 2 & 2 & 0.0027 & 1256.11 & 44\% & 20\% & 36\%\\
 2 & 4 & 0.0014 & 2554.93 & \textbf{100\%} & 0\% & 0\%\\
 2 & 8 & 0.0008 & 5153.64 & \textbf{100\%} & 0\% & 0\%\\
\hline
 4 & 1 & 1.4474 & 429.63 & 0\% & 0\% & 100\%\\
 4 & 2 & 0.0043 & 849.15 & 24\% & 0\% & 76\%\\
 4 & 4 & 0.0017 & 1733.71 & 97\% & 0\% & 3\%\\
 4 & 8 & 0.0010 & 3446.56 & 97\% & 0\% &	3\%\\ 
 \bottomrule
\end{tabular}
\label{table:train}
\end{table}

\begin{figure}[!htbp]
\centering
        \begin{subfigure}[!htbp]{0.8\linewidth}
            \centering
            \begin{tikzpicture}
                \begin{axis}
                    [
                    title={Batch Size 1},
                    xlabel={Epoch},
                    ylabel={Percentage \%},
                    xmin=0, xmax=10,
                    ymin=-25, ymax=125,
                    xtick={1, 2, 3, 4, 5, 6, 7, 8, 9},
                    ytick={0, 25, 50, 75, 100},
                    legend pos=outer north east,
                    legend cell align={left},
                    ymajorgrids=true,
                    grid style=dashed,
                    height = 0.5\linewidth,
                    width = 0.75\linewidth
                    ]
                    
                    \addplot[
                        color=green,
                        mark=square,
                    ]
                    coordinates {
                        (1,25)(2,0)(4,0)(8,95)
                    };
                    \addlegendentry{\small Success}
                    
                    \addplot[
                        color=orange,
                        mark=x,
                    ]
                    coordinates {
                        (1,30)(2,0)(4,0)(8,0)
                    };
                    \addlegendentry{\small Failure}
                    
                    \addplot[
                        color=blue,
                        mark= o,
                    ]
                    coordinates {
                        (1,45)(2,100)(4,100)(8,5)
                    };
                    \addlegendentry{\small Exception}
                \end{axis}
            \end{tikzpicture}
        \end{subfigure}
        \vskip\baselineskip
        \begin{subfigure}[!htbp]{0.8\linewidth}
            \centering
            \begin{tikzpicture}
                \begin{axis}
                    [
                    title={Batch Size 2},
                    xlabel={Epoch},
                    ylabel={Percentage \%},
                    xmin=0, xmax=10,
                    ymin=-25, ymax=125,
                    xtick={1, 2, 3, 4, 5, 6, 7, 8, 9},
                    ytick={0, 25, 50, 75, 100},
                    legend pos=outer north east,
                    legend cell align={left},
                    ymajorgrids=true,
                    grid style=dashed,
                    height = 0.5\linewidth,
                    width = 0.75\linewidth
                    ]
                    
                    \addplot[
                        color=green,
                        mark=square,
                    ]
                    coordinates {
                        (1,0)(2,45)(4,100)(8,100)
                    };
                    \addlegendentry{\small Success}
                    
                    \addplot[
                        color=orange,
                        mark=x,
                    ]
                    coordinates {
                        (1,0)(2,15)(4,0)(8,0)
                    };
                    \addlegendentry{\small Failure}
                    
                    \addplot[
                        color=blue,
                        mark=o,
                    ]
                    coordinates {
                        (1,100)(2,40)(4,0)(8,0)
                    };
                    \addlegendentry{\small Exception}
                \end{axis}
            \end{tikzpicture}
        \end{subfigure}
        \vskip\baselineskip
        \begin{subfigure}[!htbp]{0.8\linewidth}
            \centering
            \begin{tikzpicture}
                \begin{axis}
                    [
                    title={Batch Size 4},
                    xlabel={Epoch},
                    ylabel={Percentage \%},
                     xmin=0, xmax=10,
                    ymin=-25, ymax=125,
                    xtick={1, 2, 3, 4, 5, 6, 7, 8, 9},
                    ytick={0, 25, 50, 75, 100},
                    legend pos=outer north east,
                    legend cell align={left},
                    ymajorgrids=true,
                    grid style=dashed,
                    height = 0.5\linewidth,
                    width = 0.75\linewidth
                    ]
                    
                    \addplot[
                        color=green,
                        mark=square,
                    ]
                    coordinates {
                        (1,0)(2,15)(4,90)(8,95)
                    };
                    \addlegendentry{\small Success}
                    
                    \addplot[
                        color=orange,
                        mark=x,
                    ]
                    coordinates {
                        (1,0)(2,0)(4,0)(8,0)
                    };
                    \addlegendentry{\small Failure}
                    
                    \addplot[
                        color=blue,
                        mark=o,
                    ]
                    coordinates {
                        (1,100)(2,85)(4,10)(8,5)
                    };
                    \addlegendentry{\small Exception}
                \end{axis}
            \end{tikzpicture}
        \end{subfigure}
        \captionsetup{justification=centering}
        \caption{Testing Results - Conventional Job Shop Scheduling} 
        \label{fig:test}
\end{figure}

\begin{figure}[!htbp]
\centering
        \begin{subfigure}[!htbp]{0.75\linewidth}
            \includegraphics[width=\linewidth]{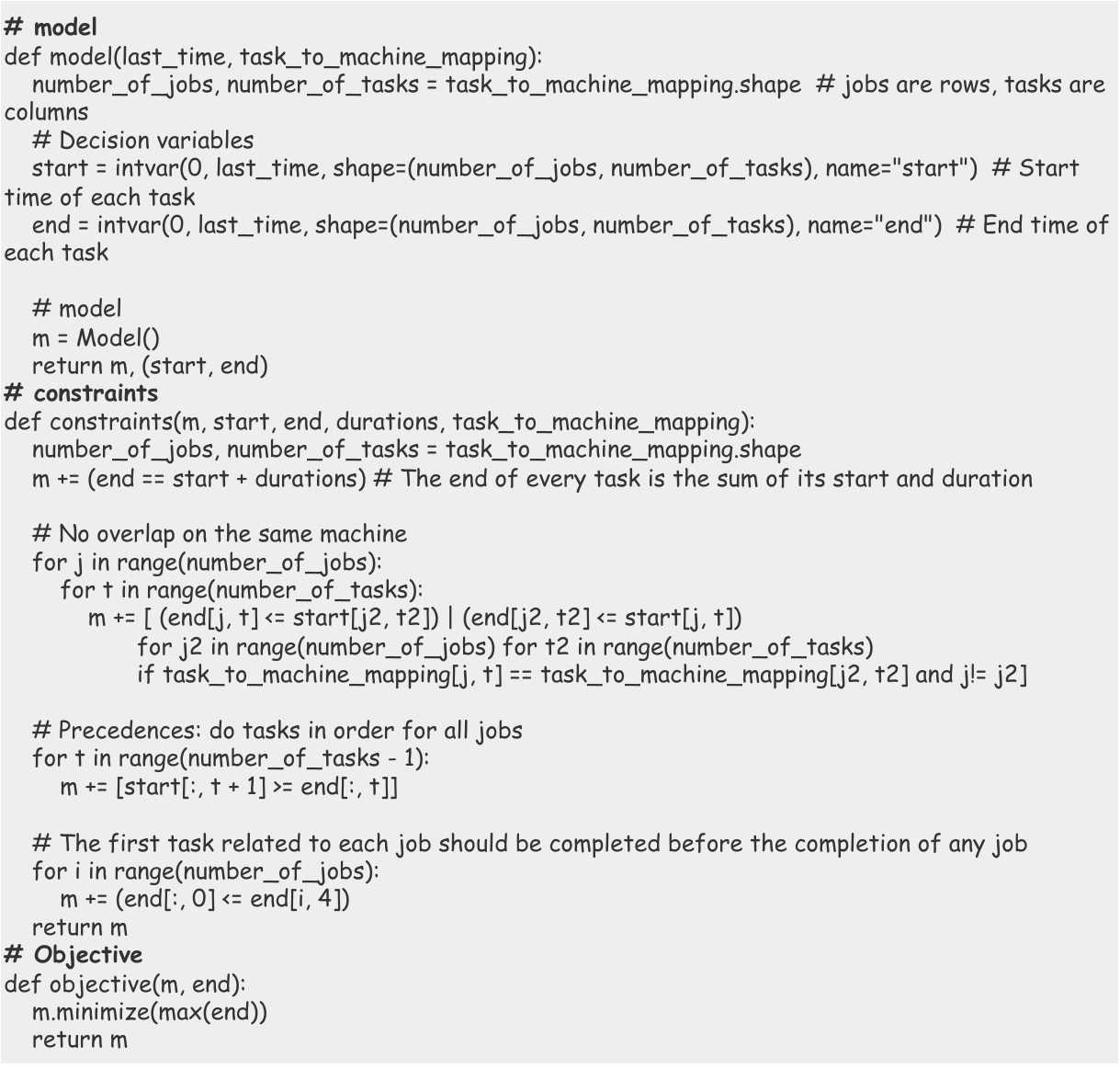}
            \caption{ Part of the generated problem formulation. Variable last\_time means the latest completion time by summing up all processing times of all jobs. Variable task\_to\_machine\_mapping contains for all the jobs, which operation should run on which machine.}   
        \end{subfigure}
        \vskip\baselineskip
        \begin{subfigure}[!htbp]{0.75\linewidth}
            \includegraphics[width=\linewidth]{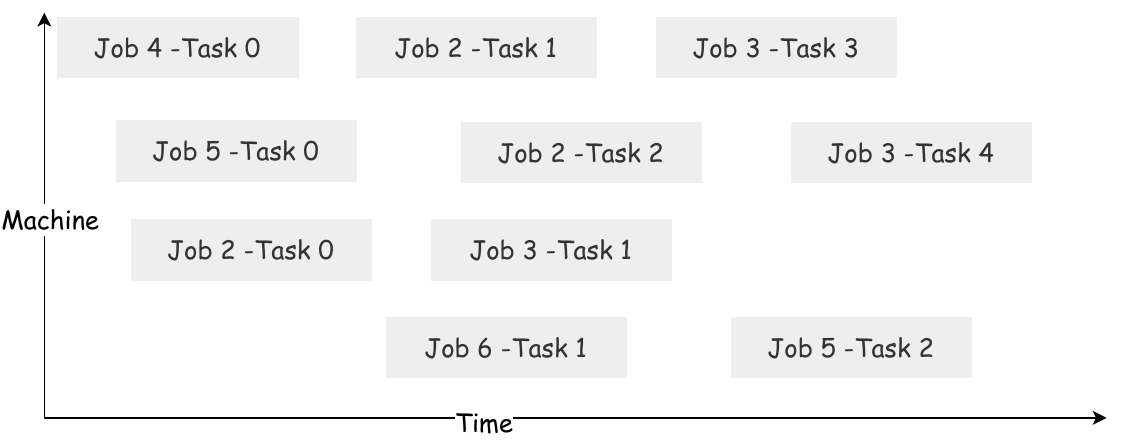}
            \caption{The output of the executed problem formulation.}    
        \end{subfigure}
        \caption{A sample problem formulation generated by our framework and the output of the executed problem formulation.} 
        \label{fig:example}
\end{figure}

We investigate vector embeddings of the fine-tuned LLM to identify how it responds to different problem descriptions. The pre-trained model we use for fine-tuning is an encoder-decoder transformer model \citep{vaswani2017attention}. The encoder embeddings show how the fine-tuned LLM captures input problem descriptions. The decoder embeddings show how the fine-tuned LLM represent the target problem formulations. However, these embeddings are $512\times1024$ dimensional vectors. Therefore, we use Principal Component Analysis (PCA) as a dimensionality reduction technique and pick the main two principal components for our visualisations (Figure~\ref{fig:vector embeddings}). Encoder embeddings are grouped into a few clusters. These clusters have problem descriptions related to different instructions (prompts). Encoder embeddings show the variety of our problem descriptions. The decoder embeddings have more clusters. However, most clusters contain problem formulations related to one type of instruction. The clusters related to decoder embeddings show the variety of problem formulations that need to be constructed. Overlapping clusters occur when there are similar tokens. In this research, similar tokens mean, identical words in problem descriptions or identical variables (syntax related to the modelling language) in problem formulations. The constructed problem formulations for this case study by the fine-tuned model can be found here \footnote{ \url{https://github.com/pivithuruthejanamarasinghe/AI-Copilot-Artifacts}.}. A sample constructed problem formulation and its executed output can be seen in Figure \ref{fig:example}.

\begin{figure}[!htbp]
\centering
        \begin{subfigure}[!htbp]{0.8\linewidth}
            \centering

        \end{subfigure}
        \captionsetup{justification=centering}
        \caption{ Vector embeddings of the fine-tuned LLM - Conventional Job Shop Scheduling} 
        \label{fig:vector embeddings}
\end{figure}

\subsection{Real-World Production Scheduling Problems}

For the experiments, we make a few assumptions. When preparing the schedules, we do not consider transportation time between machines and between the waiting area and machines. We randomly generate quantities to simulate different machine scenarios and different machine requirements. Each operation is performed on a single machine and not split among multiple machines. To our convenience, we use a quarter-hour as a one-time unit. We randomly assign due dates using a uniform distribution because we do not have access to Enterprise resource planning (ERP) data. However, using random probability distribution is not a concern as it is a common practice in the scheduling community. The distribution can be adjusted to represent different situations e.g. different levels of urgency.

This time we created fifty production scheduling problem instances and their formulations \footnote{ \url{https://github.com/pivithuruthejanamarasinghe/AI-Copilot-Data-Real-World-Scenario}.}. Similar to the previous case study we constructed problem formulations using the CPMpy \citep{guns2019increasing} library. The constructed problem formulations have token lengths in the range of 1800 to 3400. The problem descriptions have less than 600 tokens similar to the previous time. A sample problem description can be,

\begin{quote}
``
We need a schedule for XYZ Factory that focuses on minimising makespan. The job details are in the source-quantity.txt file, with each job categorised as very small, small, or large in quantity. The Little Fang Machine specialises in very small quantities, while the 7-Head Machine processes small quantities. If a job exceeds a machine's capacity, it won't be processed by that machine. Once this task is completed, please ensure the solutions are thoroughly documented and visually illustrated.
"
\end{quote}

In this case study, we use sixteen prompts as instructions. Therefore the dataset expands to 800 instances. We follow the fine-tuning process exactly similar to the previous case study. The training (Table \ref{table:train_2}) results have similar behaviour as in the previous case study. For all batch sizes, we have been able to obtain more than 95\% success rate in the training data. Furthermore, the testing (Figure \ref{fig:test_2}) results have a similar relationship as the previous case study. For all batch sizes, we have been able to obtain more than 95\% success rate in the testing dataset for this case study. The failure rate has not exceeded 25\% and the exception rate has reduced with the increase of epoch count. The batch size one and epoch count four or batch size two and epoch count four is a suitable training parameter setting for this case study as they have more than 95\% success rate and lower training time. 

The loss convergence for this case study is also similar to the previous case study (Figure \ref{appendix:cs2}). Based on the loss convergence diagrams, batch size two and epoch count four or batch size four and epoch count eight are good parameter settings for training as they have low differences between the curves, curves are not identical and curves have fluctuations. For this case study, encoder PCA embeddings are more scattered compared to the previous case study (Figure \ref{fig:vector embeddings_2}). This means problem descriptions have more variety. Furthermore, decoder PCA embeddings have more clusters compared to the previous case study. This is because we use more instructions in this case study. Similar to the previous case study we observe overlapping clusters due to similar words in problem descriptions and similar variables in problem formulations. The constructed problem formulations for this real-world case study by the fine-tuned model can be found here \footnote{ \url{https://github.com/pivithuruthejanamarasinghe/AI-Copilot-Artifacts-Real-World-Scenario}.}.

\begin{table}[!htbp]
\centering
\captionsetup{justification=centering}
\caption{Training Results - Real-World Production Scheduling}

        \end{subfigure}
        \captionsetup{justification=centering}
        \caption{Testing Results - Real-World Production Scheduling} 
        \label{fig:test_2}
\end{figure}

\begin{figure}[!htbp]
\centering
        \begin{subfigure}[!htbp]{0.8\linewidth}
            \centering
            %
        \end{subfigure}
        \captionsetup{justification=centering}
        \caption{Vector embeddings of the fine-tuned LLM - Real-World Production Scheduling} 
        \label{fig:vector embeddings_2}
\end{figure}

\subsection{Expansion to Linear Programming Datasets}

To compare our framework with other existing approaches for problem formulations, we evaluate the performance of our framework using the linear programming dataset LPWP \citep{ramamonjison2022augmenting}. This LPWP is a more refined version of the NL4Opt \citep{ramamonjison2023nl4opt} dataset. In our comparison, we compare the performance of our framework with Chain-of-Experts (CoE) \citep{xiao2023chain}, Chain of Thought (CoT) \citep{wang2022self}, Progressive-Hint Prompting (PHP) \citep{zheng2023progressive} and standard GPT. As these approaches are prompt engineering-based approaches, they have used all the problems in the LPWP dataset to measure the accuracy. However, our framework is a fine-tuning-based approach. Therefore we need a training and a validation dataset to fine-tune the pre-trained model. Then we need a test dataset to measure the performance. Therefore, similar to the previous two case studies, we randomly take 70\% of the LPWP dataset as the training dataset, 10\% as the validation dataset, and 20\% as the test dataset.

LPWP dataset has 288 problem descriptions and expected values. However, problem formulations related to the problem descriptions are not provided. Therefore, as the first step, we had to construct problem formulations. To do that we used GPT-3.5 \citep{openai2023chatgpt}. However, in some problem descriptions expected values in the LPWP dataset and the values given by the constructed problem formulations had mismatches. We manually reviewed such scenarios and resolved the conflicts. In most cases the expected values in the LPWP dataset were incorrect. Also, there were a few cases where problem formulations generated by GPT-3.5 had issues. The refined LPWP dataset by us can be found here\footnote{ \url{https://github.com/pivithuruthejanamarasinghe/LPWP}.}. To measure the performance of CoE, CoT, PHP, and standard GPT with respect to the LPWP dataset, we use the framework introduced by COE\footnote{ \url{https://github.com/xzymustbexzy/Chain-of-Experts}.}. This framework needs a commercially available LLM to construct formulations using the above four techniques. Therefore we use gpt-3.5-turbo as the LLM for this comparison.

The performance of our framework compared to the four approaches on the testing dataset can be found in Table \ref{table: comparison}. Our framework has nearly more than 30\% performance improvement over others. The four existing approaches use prompt engineering with commercially available LLMs. For smaller LLMs without adequate trained knowledge prompt engineering and in-context learning alone cannot solve complex optimisation problems. In our framework, we follow a fine-tuning approach. By fine-tuning we provide the necessary knowledge to the pre-trained LLM on constructing problem formulations. Therefore our framework performs better compared to existing solutions. However, our framework has an opportunity to further improve performance. One opportunity is to perform the problem description regeneration step as in the previous case studies to increase the number of problem descriptions. We did not do this in the LPWP dataset because we already had 288 problem descriptions. The other opportunity is to focus on capturing the wide range of words available in LPWP problem descriptions. The variety in the LPWP problem descriptions can be seen in the vector embeddings (Figure \ref{fig:vector embeddings_3}). Vector embeddings of problem descriptions are scattered compared to the previous two case studies. Especially in the encoder embeddings, we observe that only on a few occasions compared to the other two case studies, vector embeddings of two problem descriptions overlap. This means tokens contained in these problem descriptions are different from each other. In general, what LLMs do is predict the next token based on the previous tokens based on conditional probability. In the LPWP dataset, as there is a large variety of tokens, these conditional probabilities have a more flattened distribution. 

\begin{table}[!htbp]
\centering
\captionsetup{justification=centering}
\caption{Performance Comparison on the test LPWP data}
\begin{tabular}
{>{\raggedright\arraybackslash}p{0.40\linewidth}  >{\raggedleft\arraybackslash}p{0.1\linewidth} >
{\raggedleft\arraybackslash}p{0.1\linewidth}  >
{\raggedleft\arraybackslash}p{0.1\linewidth}}
\toprule
 \textbf{Method} & \textbf{Success} & \textbf{Failure} & \textbf{Exception}\\ [0.5ex] 
\midrule
Standard & 43\% & 24\% & 33\%\\
Chain of Thought & 7\% & 7\% & 86\%\\
Progressive-Hint Prompting & 2\% & 3\% & 95\%\\
Chain-of-Experts (CoE) & 43\% & 31\% & 26\%\\
Our framework & \textbf{72\%} & \textbf{26\%} & \textbf{2\%}\\
\bottomrule
\end{tabular}
\label{table: comparison}
\end{table}

\begin{figure}[!htbp]
\centering
        \begin{subfigure}[!htbp]{0.8\linewidth}
            \centering
            \begin{tikzpicture}
                \begin{axis}
                    [
                    title={PCA for embeddings of the encoder},
                    xlabel={PC1},
                    ylabel={PC2},
                    xmin=-100, xmax=100,
                    ymin=-80, ymax=100,
                    xtick={-50, 0, 50},
                    ytick={-60, -40, -20, 0, 20, 40, 60,80},
                    legend pos=outer north east,
                    legend cell align={left},
                    ymajorgrids=true,
                    grid style={line width=.1pt, draw=gray!10},
                    height = 0.65\linewidth,
                    width = 0.75\linewidth,
                    major grid style={line width=.2pt,draw=gray!30},
                    xmajorgrids=true, 
                    ymajorgrids=true
                    ]
                    
                    \addplot[
                        color=cyan,
                        mark=*,
                        only marks,
                        mark size=1pt,
                    ]
                    coordinates {
                        (57.64668638853042, 38.47559655325782)
                        (81.16157392699589, 63.401751670906094)
                        (21.460263040515496, 24.061205523077977)
                        (4.436827633071843, -37.56598347166566)
                        (-57.452054060233436, 4.093333951974732)
                        (-61.73949590724482, 26.46864310443671)
                        (-26.49610468290374, 81.48078355733843)
                        (-76.35777283909714, -11.041181674165339)
                        (-14.420449345202847, 6.497006482131731)
                        (57.6215189958904, -22.766009305231968)
                        (52.81228362173008, 6.079523514974985)
                        (-47.82036412667697, -54.87679054260381)
                        (-13.710060878280466, -28.21648998105212)
                        (-43.392259879404214, -40.99735726034901)
                        (-66.03561602532325, -3.948239048517164)
                        (-44.683996761658335, 45.9491798366038)
                        (0.7408927693216045, 34.18914335605571)
                        (-12.026918131913138, -14.48530772371724)
                        (17.38143685924909, -63.03275187483081)
                        (-11.743913769911552, 15.862557505937286)
                        (-51.09579645450761, 20.14021278547938)
                        (-32.08303751305061, -25.624558153118393)
                        (-54.60850052999201, -11.156594466233937)
                        (-41.08406960568498, 23.373328466736467)
                        (27.404553360800275, -32.43026329533898)
                        (65.06936198584208, 70.085457550065)
                        (-12.206376812243711, -34.4015114618468)
                        (36.08203593812186, -1.3619836648077277)
                        (16.114087720039464, 2.9377085743461184)
                        (59.644734506832904, -55.21757959274653)
                        (79.38796137347751, -35.24391478091724)
                        (-47.61664882173057, -7.812492591802223)
                        (31.719455345073676, -18.763808705038823)
                        (-13.797664325342824, 42.87963147554858)
                        (11.590312778653058, 21.374188650161052)
                        (-11.751686713358806, 25.18385992826065)
                        (16.391468115393874, 35.49784779723864)
                        (45.42246379255879, 9.188556551134191)
                        (39.567673645510666, -34.029509500556365)
                        (-53.233272789382625, 28.919472484705036)
                        (-5.377078863506254, -33.62955046061788)
                        (-0.04673166426369253, 47.044562108430746)
                        (40.57144152201715, 3.7697426063258628)
                        (32.0667452730071, -32.46282802129866)
                        (37.039464384188825, -57.59498569900266)
                        (-49.31911064019465, -6.744060650292923)
                        (6.244826061402928, -0.4210958782696175)
                        (12.1418484343872, -16.707212030749805)
                        (27.01997678672798, -2.8410253380209975)
                        (35.98124031919416, -2.059504017780715)
                        (-29.234605132581667, -29.85111098120157)
                        (-34.290906173394696, 8.511604246933656)
                        (73.18677372283051, 69.36034272453826)
                        (50.36191566208061, -32.130891159825744)
                        (75.0541858649327, 12.716394040528542)
                        (54.51663563927095, 1.0782840639239284)
                        (-30.82339021043664, -15.981271061016074)
                        (-0.9706581537822982, 13.67087670047929)
                        (-28.727395938616468, -52.45240492614397)
                        (-30.457737124259413, -43.32687651939396)
                        (51.691294315341814, 11.40058445137606)
                        (14.58741979338231, -32.54278104370359)
                        (36.14459669921067, 5.445802734074607)
                        (-36.44915655138475, 29.11133733236677)
                        (51.4050094383222, -38.255629326270046)
                        (42.406754089345334, -48.56331995371697)
                        (71.20549483577338, -19.00644844108995)
                        (27.300128825909336, 0.8936216238639334)
                        (-58.3343435286487, 44.009416923224805)
                        (-10.098535601228916, 16.89597756922487)
                        (-59.09799198592539, -24.423305507749458)
                        (-36.05974685161813, -20.236535796764166)
                        (64.87315004243585, 58.67245850758571)
                        (-2.927923145151302, -8.486922716009884)
                        (14.615199852872687, -51.96398542993643)
                        (34.87935483259697, 16.051281322294287)
                        (-54.206406598310565, -1.7914447861519989)
                        (7.834812396851021, 34.82238559340913)
                        (32.97648004929592, -2.4366795986798553)
                        (-2.918654238461376, -21.174513618581738)
                        (5.084375746727143, 10.929732429458516)
                        (26.364279120077335, 33.83323904888588)
                        (-42.46876729101986, -4.407422822128813)
                        (-16.004878550760026, 2.215248313733218)
                        (-38.661365147683185, 14.67656375464897)
                        (10.804637890454575, 0.550460565876899)
                        (59.314439966304306, 61.543094339162984)
                        (-24.149387911557817, -34.728048771315606)
                        (-36.483518186909784, -6.25580214571667)
                        (-16.48739484864387, 13.479531237144268)
                        (35.79156402694032, 22.530356417194117)
                        (-30.409902254318567, 63.918645585691586)
                        (-39.45034925565223, 58.13134609955364)
                        (-13.640596157211203, 35.84455346069538)
                        (3.315208042187936, 28.057742224156403)
                        (-33.22546392629512, 42.198637285425576)
                        (75.43884750549148, 30.631065017061783)
                        (51.41526705297877, 50.021109265127315)
                        (72.19087703219374, -2.219303020284429)
                        (-57.63343103333777, -39.456236179196246)
                        (22.370377082675002, 67.70275567235672)
                        (-57.9700049404941, 40.850107774009324)
                        (-31.485517635258084, -24.99519163807911)
                        (-11.837794164513337, 7.36502703608511)
                        (-19.113738236818033, -15.916405859068908)
                        (-28.324470882442434, 12.822879571028855)
                        (-31.04653645726759, 47.7573154991151)
                        (-28.118348583062957, 19.21471173775105)
                        (-40.3570295164172, 16.124023733350203)
                        (77.19458062529563, 47.05869008488695)
                        (38.73563621932756, 80.69200735618985)
                        (-47.50343299794565, -10.076551911653375)
                        (-44.09672165928414, -5.7339017427527255)
                        (64.63065445694838, 14.68677620858)
                        (-37.73941600653543, 26.69275462709081)
                        (-19.668046682544528, -50.90613720299213)
                        (-24.10309268773409, 16.818646698229347)
                        (21.76505804856892, -65.44677690849755)
                        (79.61414742581415, 61.98023995887799)
                        (-40.21336811836481, 18.149566541291666)
                        (13.61192356220625, -28.93204367086347)
                        (-40.70142821851437, -11.131033962984455)
                        (11.20977307059286, 6.085926313242507)
                        (-3.465682726934646, -26.62703225787319)
                        (-58.32066573962407, 42.7014980311835)
                        (-66.13787544869831, 10.116870045162907)
                        (-21.49632774806337, 0.6976093420752577)
                        (24.93485578397857, -41.68193197414385)
                        (-15.611784038312477, -39.486145971604614)
                        (-61.360462870645065, -5.613673190891287)
                        (41.60048330554674, -19.122387194252454)
                        (-2.835161520329335, -4.543077954121921)
                        (-70.70511242094987, -18.46664522715468)
                        (-40.325922267698694, 4.5355963431120845)
                        (10.834994561944399, 22.747153805504247)
                        (-34.466675566216075, -37.514679030826485)
                        (-12.341103602059034, -37.20416239426781)
                        (43.03928590646408, -30.653833757080157)
                        (23.011421482514535, -53.85189644347614)
                        (-22.767600855012073, -47.01653695150188)
                        (-31.72797157829757, -8.49986492725004)
                        (55.090492133808425, -2.2246062669793463)
                        (34.371803745560975, 17.760993843207626)
                        (30.805124318653842, -33.476906844335375)
                        (-45.53024123530514, 78.01332731820308)
                        (51.53717172162597, -72.05437229916879)
                        (12.46344925605253, -11.390051635085923)
                        (-44.28635633789172, 17.2013101030294)
                        (75.6765731010246, -20.148573670648823)
                        (-43.118069276780645, 13.070172955882084)
                        (-15.19107949756374, -19.681543808199624)
                        (-29.01778881523265, 12.178534272893028)
                        (25.72130274373848, -38.782272599843566)
                        (67.49157917379996, -0.7754796942468392)
                        (47.08654003265363, -37.30087079920625)
                        (-3.56241757034212, -53.13620142671543)
                        (38.57841553616097, -13.06351688543385)
                        (-77.50572751909837, 28.538855927857405)
                        (-35.841117610314186, -28.725385220906833)
                        (-22.81280349990517, -3.493705135266488)
                        (21.410170905354676, 16.264061428225148)
                        (-61.14668165850851, 54.58303911769218)
                        (13.158280113924022, -17.583708159031243)
                        (-11.73706808823408, -13.076276559304485)
                        (63.80959509004838, -9.134502132042336)
                        (-59.526293067208876, -31.985216155423217)
                        (-65.6292177841527, 35.5056025952906)
                        (39.7905148057104, -6.4310772219965795)
                        (-29.644941843043867, -59.84862689049227)
                        (-46.62453816151759, -34.2522500612553)
                        (6.641402896954848, 1.7254167406774623)
                        (7.386055609890567, -20.390351137526576)
                        (63.18080421884231, 19.15823983376044)
                        (-57.53894000460853, 0.44524606457009824)
                        (-48.43472184475914, -18.5059673229133)
                        (62.22043250978872, -27.28516060795057)
                        (10.24091050149977, -0.1572055494481455)
                        (48.66861262194391, 8.782798061912859)
                        (70.46374343061518, 17.44356006256493)
                        (38.75524489689571, 38.12000502413768)
                        (29.845300878410498, 23.915850856352375)
                        (22.968263291397413, -14.00491819120427)
                        (-3.3178133916659465, -16.983779203564378)
                        (22.012295663037634, 41.61975449992102)
                        (-13.533505199921496, 31.332463080003873)
                        (-3.907960707097818, 9.895271826396929)
                        (-54.22693362566002, -42.61388737718794)
                        (-41.09216532782889, 26.1579829057818)
                        (35.54256563272337, -24.24802671931772)
                        (-25.02046785518771, 64.74295261880113)
                        (70.25189953080074, -25.084276429462076)
                        (-50.12726528425991, -14.20312983479213)
                        (20.143093300713694, 31.678092269060297)
                        (-33.72157905162998, -28.61102991843951)
                        (4.021595680514702, 11.172827939462568)
                        (67.52790786426641, -23.594815510650786)
                        (58.45549296284765, -29.190950663989323)
                        (-41.84727977668534, 24.951555380816806)
                        (1.0396752047963238, -50.73676683147727)
                        (-57.89047447525856, 3.21143840793823)
                        (-67.40310931027602, -19.705654496034292)
                    };
                   
                \end{axis}
            \end{tikzpicture}
        \end{subfigure}
        \vskip\baselineskip
        \begin{subfigure}[!htbp]{0.8\linewidth}
            \centering
            \begin{tikzpicture}
                \begin{axis}
                    [
                    title={PCA for embeddings of the decoder},
                    xlabel={PC1},
                    ylabel={PC2},
                    xmin=-1000, xmax=1500,
                    ymin=-1000, ymax=1500,
                    xtick={-1000, -500, 0, 500, 1000},
                    ytick={ -500, 0, 500, 1000, 1500},
                    legend pos=outer north east,
                    legend cell align={left},
                    ymajorgrids=true,
                    grid style={line width=.1pt, draw=gray!10},
                    height = 0.65\linewidth,
                    width = 0.75\linewidth,
                    major grid style={line width=.2pt,draw=gray!30},
                    xmajorgrids=true, 
                    ymajorgrids=true
                    ]
                    
                    \addplot[
                        color=cyan,
                        mark=*,
                        only marks,
                        mark size=1pt,
                    ]
                    coordinates {
                        (-651.217459610322, -411.1393794088271)
                        (874.8781035555643, -316.9153053949802)
                        (-535.8734338164121, 485.8359694720983)
                        (1359.1782393755493, -550.7859587738857)
                        (552.0488527150619, -213.72231615731505)
                        (-122.0933604189896, 858.2449976537954)
                        (-776.8613437618802, -587.7304076442571)
                        (-795.9218479765821, -686.2286687810923)
                        (590.010378239067, -124.02509613496866)
                        (1333.4422506580952, -519.1874283372943)
                        (1085.6723944499734, -383.8963449200123)
                        (466.3310811969411, -141.28937330461642)
                        (-855.2913382109297, 29.795031644873788)
                        (634.2005772281266, -210.49770341821545)
                        (740.8741479704844, 31.716729290459313)
                        (304.6216159194117, -408.8715149730898)
                        (-327.7809889351547, 849.2510131673688)
                        (-79.11175250702749, 680.4677751207965)
                        (-845.0565964992423, -660.6269897790614)
                        (-334.8735061074844, 681.0520001878695)
                        (840.8273936345907, -73.32035495101945)
                        (536.6929249580519, -17.288913307531562)
                        (1021.1498360140341, -241.3295796203522)
                        (630.0521732379764, -299.00674518965553)
                        (-699.1822730940174, -337.662048147198)
                        (332.55727034070816, 493.28222543993076)
                        (-808.8214632285109, -638.2169485021972)
                        (-360.56278850571346, 480.84257921404634)
                        (1043.4774003009074, -339.2462548628807)
                        (-723.8793054568079, -300.44421294592377)
                        (543.6769573465883, 665.8152973008586)
                        (203.59787825705698, -408.7507604524581)
                        (-595.2136443812058, 178.6591261369781)
                        (-621.5744220816289, 61.90447268040454)
                        (-525.9211320236248, 358.03963480912796)
                        (1215.6822920487286, -458.80247921622663)
                        (1197.414226638823, -361.38861794771054)
                        (-354.86648700123123, 480.2533377064228)
                        (-293.3411669614627, 970.8265621902793)
                        (814.8882353654338, -223.95273029273562)
                        (-76.18160012721198, 718.3748375683831)
                        (-259.03934176523006, 642.5152631193453)
                        (-245.6612687570522, 201.4605191559402)
                        (-372.86270263675, 18.55542843360559)
                        (654.7342452580424, -216.86393645109044)
                        (747.1505085193878, -309.7424701270944)
                        (528.6783484983442, 984.0275479379935)
                        (-145.32061859999587, 562.5199885263444)
                        (899.3897549131337, -150.69666025725093)
                        (1260.0166204431084, -442.76559872257246)
                        (-315.5656626543557, 722.4097220543033)
                        (83.81098409570536, 445.3339249670931)
                        (-496.8179283549076, -383.5570694120603)
                        (-800.9803461519423, -143.20916903706666)
                        (1239.919341585172, -424.1772981570411)
                        (-601.317542058694, -91.5264096212947)
                        (-809.9807719035726, -694.3481754709342)
                        (644.8835769786317, -153.15790793267394)
                        (-14.197593268119935, 780.8661572289831)
                        (-838.7706321755365, -204.41558820030713)
                        (91.17905550003718, 723.0777251245992)
                        (-708.8228322432423, -495.44124785823647)
                        (660.899076920281, 488.07070572910663)
                        (180.8893931426694, -150.0959195150393)
                        (964.7163735163743, -291.6332581910373)
                        (-452.38809172908697, 390.4132522573675)
                        (1162.9385817555533, -362.19813430011084)
                        (-786.2497728420364, -469.06186719179306)
                        (834.0172347957671, -209.2176753815385)
                        (908.8665738682021, -232.27991990588856)
                        (-772.0617115361383, -707.0332728527849)
                        (-685.0205074724688, 236.50085648600586)
                        (-198.24772129001164, 603.9645374347025)
                        (-355.97639247298656, 600.1870785460187)
                        (997.3141410602483, -272.2523889030126)
                        (-765.4871052977418, -567.4196994927759)
                        (511.1259508980304, -126.93541893042111)
                        (83.39206708234259, 945.668392274451)
                        (-691.7817123461638, 364.61207918653264)
                        (59.142061429285384, 895.0576777703664)
                        (484.7956936677311, 33.430692567562176)
                        (1182.5983378134463, -351.6306028335772)
                        (-337.3707528981895, 767.7098314136093)
                        (601.6637264285198, 657.2349575460898)
                        (-889.1559956815945, -49.24423639052232)
                        (-198.82808311622523, 693.9004838054734)
                        (408.41635804985725, 691.8888153584147)
                        (324.6268124771245, -78.08743517636441)
                        (501.6943634705169, -223.48138315125456)
                        (-777.4603999742652, 159.16750638915607)
                        (219.6540581598979, 279.8844570409124)
                        (731.52615678852, -289.76137708248467)
                        (577.1254474383117, -159.8831221306655)
                        (-559.3447343070418, 270.8890050191272)
                        (-2.755654760091451, 657.5374959816627)
                        (-913.081232555569, -160.11009115903644)
                        (-396.3519957992313, -196.41766242167236)
                        (-503.61373833702453, -222.9910740389553)
                        (771.7179109388921, -179.53261743960138)
                        (-425.7376077412325, 362.199810336894)
                        (277.2914556803617, 79.77767885095388)
                        (704.7045958812977, -113.27938453698411)
                        (-685.1125983016776, -444.2717479506346)
                        (-774.9933953020713, -595.3895762357745)
                        (269.5748663648739, -369.4961736470629)
                        (1048.4480301878107, -285.7687981691236)
                        (-516.1300180102842, 538.4178787289022)
                        (-328.3425244089991, 450.22238459213486)
                        (601.7887392741613, -170.19500402344067)
                        (256.33697561298, 879.9109007902207)
                        (-207.70851608874838, 663.2133578625918)
                        (-819.6565741134314, -26.02675996674201)
                        (-942.2815911482543, -197.89983093673732)
                        (-762.8481418496661, -606.1858186161332)
                        (845.2707054430886, -39.340871775510266)
                        (543.4459676006635, -255.81744407163683)
                        (-225.8811157914733, 254.77108170926277)
                        (-617.2586148484104, 126.42668488261968)
                        (-443.3926866242683, -279.11555473781755)
                        (862.0362424087099, -224.29042583423487)
                        (810.851990963972, -225.95353132412498)
                        (-680.183697108128, -398.8135148399908)
                        (-798.26414824045, -708.6295495331132)
                        (-314.36959158374935, 310.9435010484622)
                        (-764.4396428202716, -666.355115329637)
                        (158.69048203819568, 79.54023466774703)
                        (-631.9516338214459, 200.2931661756385)
                        (39.30133408763132, 1013.8438076876122)
                        (-767.3205908503702, -613.0672879247804)
                        (771.8452806582671, -128.43058126391824)
                        (-404.8120214368673, 242.90205812489333)
                        (1059.9539593359018, -355.34036215974044)
                        (-757.138399152416, -677.6976828069637)
                        (447.013630709426, -230.90411792937286)
                        (-761.2725214101837, -584.5853302684654)
                        (-914.6622946136116, -256.5909286088315)
                        (-572.3773264493199, -132.0940801483754)
                        (-785.8832701650211, -713.0755708898342)
                        (-213.28561536236467, -82.28068945517609)
                        (408.8481131800321, -118.26663499405814)
                        (-795.6242912717967, -153.35518676333456)
                        (-687.5592374738906, 35.205369195765904)
                        (-227.82726631524463, 334.646594826849)
                        (7.075071673858941, 433.10128489416604)
                        (-839.4240744625516, -675.7626140030145)
                        (616.3714588608793, -232.11833862383105)
                        (-646.9208550808316, -353.7118257520781)
                        (-366.2165940921946, 412.35110017794324)
                        (710.2164439024548, 719.9324495978009)
                        (889.1492511549162, -282.9043714467788)
                        (-170.4125124998131, 184.9702272151938)
                        (959.4302353710453, -212.72202342212248)
                        (298.9978333497009, 418.3008169257095)
                        (873.9533949350429, -265.24624122287764)
                        (-527.781798069527, -295.29659846727037)
                        (-531.9799115654048, 646.0614750805569)
                        (820.323054102621, -230.6340101938916)
                        (-781.0902857370564, -276.7593097482769)
                        (86.10256395105485, 823.5101101261873)
                        (-354.47709267931367, 688.6126887388046)
                        (750.652453077459, -274.5731617545004)
                        (125.62465730882006, -494.2706507470796)
                        (-109.49802145137033, 374.9315948250159)
                        (760.8290299454243, -161.97809673150596)
                        (476.7225070261524, 529.4246524034901)
                        (320.0741393480553, -332.4659371879654)
                        (-493.1394381055153, 419.25795704923763)
                        (-327.514751817587, 823.6657874566522)
                        (-774.2365912953198, -688.8806775542134)
                        (441.6409692674053, -247.18116389933974)
                        (-782.7003460381047, -654.5470415586504)
                        (160.50572683821719, 346.4102485542635)
                        (51.503435110809555, 652.080803537124)
                        (958.8111573865768, -325.4749526581786)
                        (-790.690078674097, -163.00715442545828)
                        (-755.1899123418206, 9.902409856285292)
                        (-747.4897932094511, -630.358338803843)
                        (-642.5101742221774, -461.57399143221187)
                        (-660.7638330738746, -493.78352266151103)
                        (367.83838729661727, 696.4790144725791)
                        (-408.8960297042986, 293.7790893369189)
                        (-701.940838248579, -517.0573865030042)
                        (-534.6177951577183, 91.32090316794826)
                        (-84.22425325949035, 7.711757533313068)
                        (890.0368038415784, -253.24238925506856)
                        (680.526991689365, -149.40802225621076)
                        (-759.9635544372802, -581.4167625959057)
                        (444.9755957880727, -379.1164498396173)
                        (91.05731881999219, 465.21842986847093)
                        (43.70902911852551, 812.8292285413219)
                        (-632.4104212514686, -457.6329257071488)
                        (956.7833475367664, -268.3664036298923)
                        (37.3155957335684, -64.27211849742795)
                        (1276.4035988475512, -453.9521873359168)
                        (-719.3987267169508, -487.60334064106235)
                        (49.383753903377375, 721.976170902361)
                        (526.3968423694055, 744.9135970723302)
                        (-369.0124559725519, 315.20858376156616)
                        (-802.0865414959503, -554.9165697414504)
                        (-412.2108412840746, 144.25516801913412)
                        (-584.7187939882514, 306.48120370513516)
                    };
                \end{axis}
            \end{tikzpicture}
        \end{subfigure}
        \captionsetup{justification=centering}
        \caption{Vector embeddings of the fine-tuned LLM on the LPWP Dataset} 
        \label{fig:vector embeddings_3}
\end{figure}

\section{Conclusion}
\label{conclusion}

In practice constructing a problem formulation is a time-consuming task that requires expertise. There are several prompt engineering methods in literature to automate this process with commercial LLMs. Although initial results are encouraging, the accuracy of formulations generated by these existing methods can still be significantly improved. We introduce our framework as a method to fine-tune cost-efficient LLMs to automatically construct problem formulations for optimization problems. The experiments with our two case studies demonstrate our framework's capability to accurately generate formulations for conventional and real-world production scheduling problems. Furthermore, we expanded our experiments with a linear programming dataset to compare the performance of our framework with other existing methods. Based on the results, our framework has a very competitive performance with existing methods. A major goal of this research was to leverage cost-efficient LLMs to automate this process. With prompt engineering and modularisation, we have been able to leverage cost-efficient LLMs for complex real-world optimisation problems. The datasets relevant to our case studies and comparisons are publicly available for future use. As further improvements, we will direct our research towards leveraging domain knowledge and human feedback to improve the performance of our framework.

\appendix

\section{Job Shop Scheduling}
\label{appendix:JSS}

For the static JSS problem instance, the shop, such as, the working or manufacturing environment includes a set of $M$ machines and $N$ jobs that need to be scheduled. Each job $j$ has its pre-determined route through a sequence of machines to follow and its own processing time at each machine it visits. The following notation is used to define the mathematical model for the JSS \citep{nguyen2021genetic}.

\noindent Parameters:
\begin{itemize}
    \item $J=\{1,....,j,....,N\}$:  the set of all jobs
    \item $n_j$: the number of operations of job $j$
    \item $route_j = (m_{j1},....,m_{jn_j})$:  the sequence of machines  that job $j$ will visit, where $m_{ji}$ is the machine that processes the $i^{th}$ operation of job $j$
    \item $time_j = (p_{j1},....,p_{jn_j})$: the processing times of all
operations of job $j$, where $p_{ij}$ is the processing time of the $i^{th}$ operation of job $j$
\item $r_j$: the release time of job $j$
\item $d_j$: the due date of job $j$
\item $w_j$: the weight of job $j$
\end{itemize}
Variables:
\begin{itemize}
    \item $s_{ji}$: the starting time of the $i^{th}$ operation of job $j$
    \item $e_{ji}$: the ending time of the  $i^{th}$ operation of job $j$
    \item $C_j$: the completion time of job $j$
    \item $T_j$: the tardiness of job $j$ calculated by $T_j = \max(C_j-d_j,0)$
\end{itemize}
\noindent The constraint programming formulation for the JSS  problem can be defined as follows.
\begin{align}
    \label{eq:1} \forall j \in J &: s_{j1}>r_j\\
    \label{eq:2}\forall j \in J, i\in \{1,...,n_j\} &: e_{ji} = s_{ji} + p_{ji} \\
    \label{eq:3}\forall j \in J &: C_j=e_{jn_j}\\
    \label{eq:4} \forall j \in J &: T_j=\max(C_j-d_j,0)
\end{align}
where \eqref{eq:1}: starting time of the first operation of the job should be greater than the release time of the job, \eqref{eq:2}: ending time of an operation equals to the sum of starting time and processing time of an operation, \eqref{eq:3}: completion time of a job equals to the ending time of the last operation of the job, \eqref{eq:4}: tardiness of a job equals to the difference between the job completion time and the due date of the job if it is positive or zero otherwise.

\noindent To ensure no overlap between operations (or disjunctive constraints) on the same machine:
\begin{multline}
    \forall j, k\in J, u\in \{1,...,n_j\}, v\in \{1,...,n_k\}, m\in route_j, o\in route_k :\\
    m_{ju} = o_{kv} \Rightarrow s_{ju} \geq e_{kv} \vee s_{kv} \geq e_{ju}
\end{multline}
That is if operations $u$ and $v$ from different jobs are to execute on the same machine $m_{ju}=o_{kv}$, the start time of one of these jobs must be greater than the end time of the other job.\\
\noindent There are a number of precedence constraints between the operations of a job:
\begin{align}
    \forall j\in J, i\in \{1,...,n_j-1\}: s_{j,i+1} \geq e_{ji}
\end{align}
The objective functions are defined as follows:
\begin{itemize}
    \item Makespan: Defined variable $C_{\max}$ which represents the latest completion time of any job. The objective is to minimise $C_{\max}$ subject also to constrain \eqref{eq:constrain}:
    \begin{align}
        \min C_{\max}\\
        \label{eq:constrain}\forall j\in J : C_{\max} \geq e_{jn_j}
    \end{align}
    \item Maximum tardiness: Defined variable $T_{\max}$ which represents the maximum tardiness of any job. The objective is to minimise $T_{\max}$ subject to constrain \eqref{eq:constrain_2}:
    \begin{align}
        \min T_{\max}\\
        \label{eq:constrain_2}\forall j\in J : T_{max} \geq T_j
    \end{align}
    \item Total Weighted Tardiness (TWT): The objective is to minimise cumulative tardiness’ across all jobs:
    \begin{align}
        \min \sum_{j\in J} w_jT_j
    \end{align}
\end{itemize}

\section{Flexible Job Shop Scheduling}
\label{appendix:FJSS}

Flexible Job Shop Scheduling (FJSS) is an extension of JSS. For each given operation, there is at least one instance of machine type necessary to perform it. FJSS consists of a sub-problem of assigning a machine to an operation out of a set of capable machines. Then there is the scheduling problem similar to JSS. The following notation is used to define the mathematical model for the FJSS \citep{fattahi2007mathematical}.

\noindent Parameters:

\begin{itemize}
    \item $i = 1,...,m$
    \item $j = 1,...,n$
    \item $h = 1,...,h_j$
    \item $n$ Number of jobs
    \item $m$ Number of machines
    \item $O_{j,h}$: $h^{th}$ operation of job $j$
    \item \[a_{i,j,h} = \begin{cases} 1 & \text{if $O_{j,h}$ can be performed on machine $i$}\\ 0 & \text{otherwise}\end{cases}\]
    \item $p_{i,j,h}$ Processing time of operation $O_{j,h}$ if performed on machine $i(p_{i,j,h}>0)$
    \item $L$ : A large number
\end{itemize}

Variables:

\begin{itemize}
    \item $C_{max}$ : Makespan or maximal completion time of jobs
    \item \[ y_{i,j,h} = \begin{cases}
        1 & \text{ if $i$ is selected for operation $O_{j,h}$}\\
        0 & \text{otherwise}
    \end{cases}\]
    \item \[ x_{i,j,h,k} = \begin{cases}
        1 & \text{if $O_{j,h}$ is performed on $i$ in priority $k$}\\
        0 & \text{otherwise}
    \end{cases}
    \]
    \item $t_{j,h}$ : Start time of the processing of operation $O_{j,h}$
    \item $Tm_{i,k}$ : Start of working time for machine $i$ in priority $k$
    \item $k_i$ : The number of assigned operations to machine $i$
    \item $Ps_{j,h}$ : Processing time of operation $O_{j,h}$ after select a
machine
\end{itemize}

Objective:

\begin{align}
    \min C_{max}\\
    \forall j\in \{1,...,n\} : C_{\max} \geq t_{j,h_j} +  Ps_{j,h_j}
\end{align}

Constraints:
\begin{align}
    \forall j\in \{1,...,n\} ,  h\in \{1,...,h_j\} : \sum_i y_{i,j,h}.p_{i,j,h} = Ps_{j,h}\\
    \forall j\in \{1,...,n\} ,  h\in \{1,...,h_{j-1}\} : t_{j,h} + Ps_{j,h} \leq t_{j,h+1}
\end{align}
\begin{align}
    \forall i\in \{1,..,m\} , j \in \{1,...,n\},
    h \in \{1,...,h_j\}, k \in \{1,...,k_{i-1}\}:
    Tm_{i,k} + Ps_{j,h}.x_{i,j,h,k} \leq Tm_{i,k+1}
\end{align}
\begin{align}
   \forall i\in \{1,...,m\} , j \in \{1,...,n\},
   h \in \{1,...,h_j\}, k \in \{1,...,k_{i}\}:
   Tm_{i,k} \leq t_{j,h} + (1-x_{i,j,h,k}).L
\end{align}
\begin{align}
    \forall i\in \{1,...,m\} , j \in \{1,...,n\},
    h \in \{1,...,h_j\}, k \in \{1,...,k_{i}\}:
    Tm_{i,k} + (1-x_{i,j,h,k}).L \geq t_{j,h}
\end{align}
\begin{align}
    \forall i\in \{1,...,m\} , j \in \{1,...,n\}, h \in \{1,...,h_j\}:
    y_{i,j,h} \leq a_{i,j,h}
\end{align}
\begin{align}
    \forall i\in \{1,...,m\} , k \in \{1,...,k_i\}:
    \sum_j \sum_h x_{i,j,h,k} = 1
\end{align}
\begin{align}
    \forall j\in \{1,...,n\}, h \in \{1,...,h_j\}:
    \sum_i y_{i,j,h} = 1
\end{align}
\begin{align}
    \forall i\in \{1,...,m\} , j \in \{1,...,n\} , h \in \{1,...,h_j\}:
    \sum_k x_{i,j,h,k} = y_{i,j,h}
\end{align}
\begin{align}
    \forall j \in \{1,...,n\}, h \in \{1,...,h_j\} : t_{j,h} \geq 0 \\
    \forall j \in \{1,...,n\}, h \in \{1,...,h_j\} : Ps_{j,h} \geq 0 \\
    \forall i \in \{1,...,m\}, k \in \{1,...,k_j\} : Tm_{i,k} \geq 0
\end{align}
\begin{align}
    \forall i \in \{1,...,m\}, j \in \{1,...,n\},
    h \in \{1,...,h_j\}, k \in \{1,...,k_i\}: 
    x_{i,j,h,k} \in \{0,1\}
\end{align}
\begin{align}
    \forall i \in \{1,...,m\}, j \in \{1,...,n\}, h \in \{1,...,h_j\}:
    y_{i,j,h} \in \{0,1\}
\end{align}

\begin{figure*}[!htbp]
\centering
        \begin{subfigure}[!htbp]{0.24\textwidth}
            \centering
            \includegraphics[width=\textwidth]{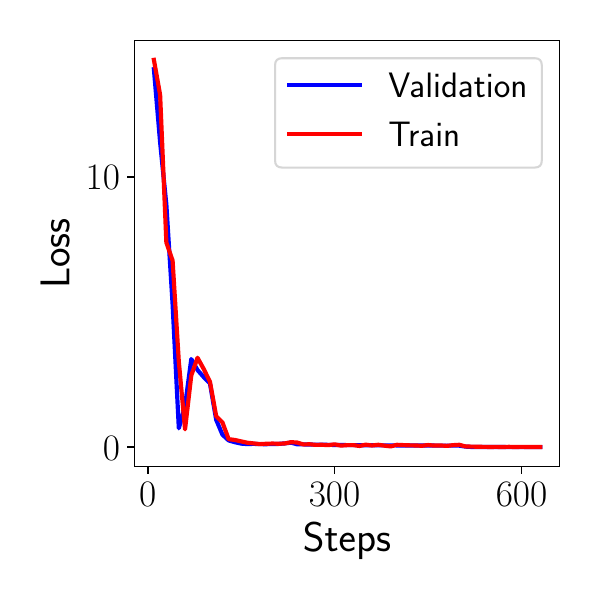}
            \captionsetup{justification=centering}
            \caption{ Batch size 1 \& epoch 1}    
        \end{subfigure}
        \hfill
        \begin{subfigure}[!htbp]{0.24\textwidth}  
            \centering 
            \includegraphics[width=\textwidth]{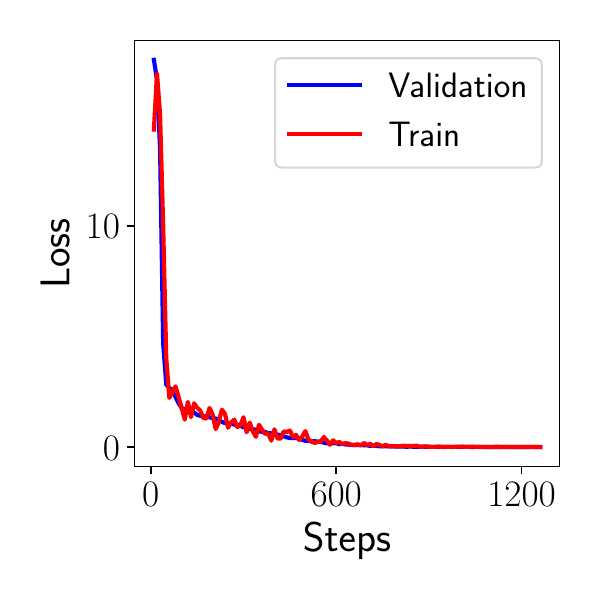}
            \captionsetup{justification=centering}
            \caption{Batch size 1 \& epoch 2}   
        \end{subfigure}
        \hfill
        \begin{subfigure}[!htbp]{0.24\textwidth}   
            \centering 
            \includegraphics[width=\textwidth]{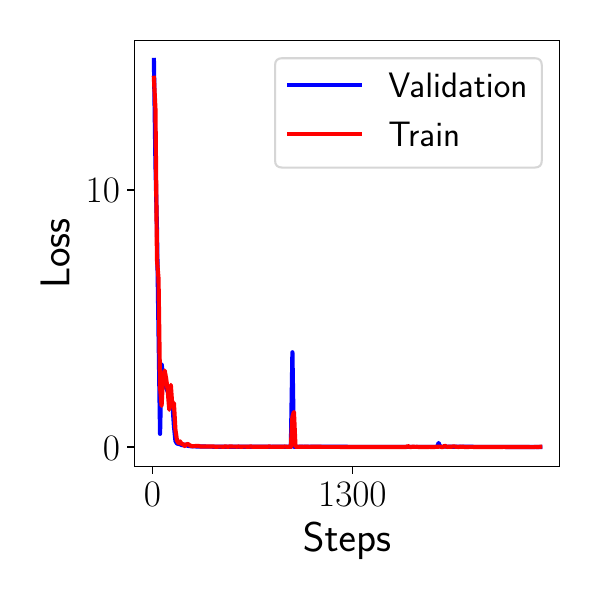}
            \captionsetup{justification=centering}
            \caption{Batch size 1 \& epoch 4}   
        \end{subfigure}
        \hfill
         \begin{subfigure}[!htbp]{0.24\textwidth}   
            \centering 
            \includegraphics[width=\textwidth]{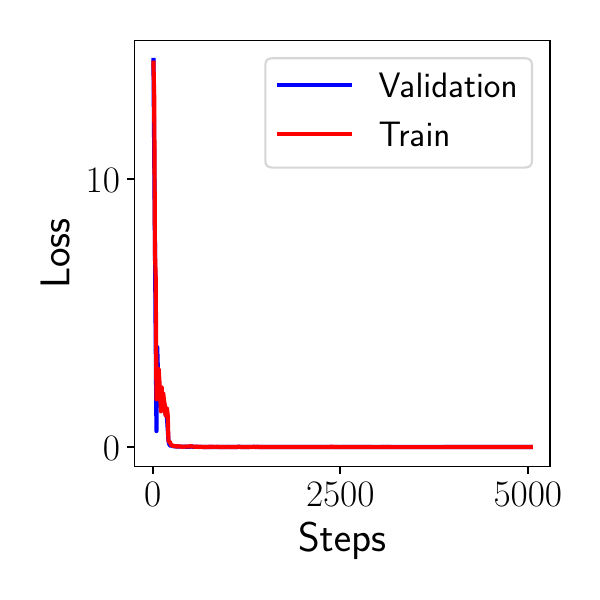}
            \captionsetup{justification=centering}
            \caption{Batch size 1 \& epoch 8}    
        \end{subfigure}
        \vskip\baselineskip
        \begin{subfigure}[!htbp]{0.24\textwidth}
            \centering
            \includegraphics[width=\textwidth]{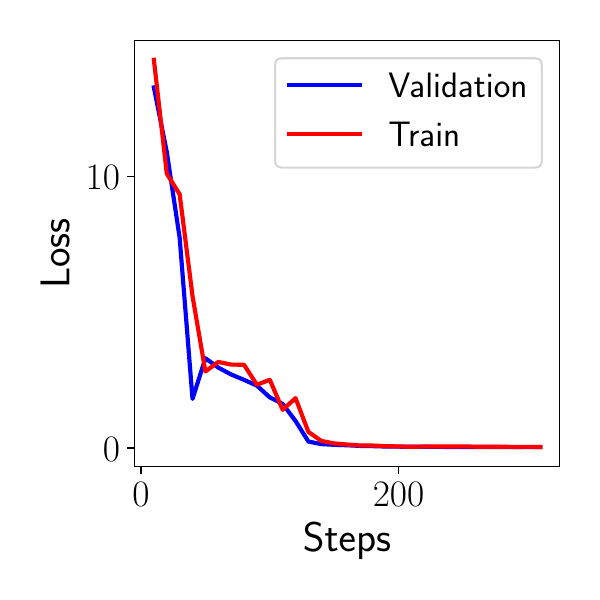}
            \captionsetup{justification=centering}
            \caption{Batch size 2 \& epoch 1}    
        \end{subfigure}
        \hfill
        \begin{subfigure}[!htbp]{0.24\textwidth}
            \centering
            \includegraphics[width=\textwidth]{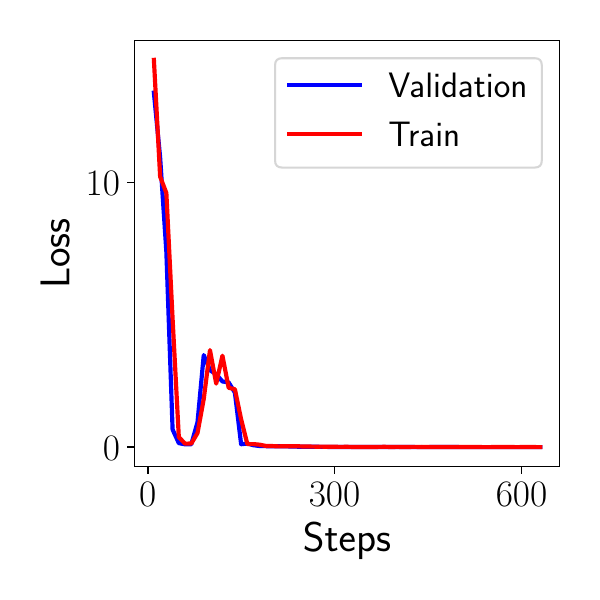}
            \captionsetup{justification=centering}
            \caption{Batch size 2 \& epoch 2}    
        \end{subfigure}
        \hfill
         \begin{subfigure}[!htbp]{0.24\textwidth}  
            \centering 
            \includegraphics[width=\textwidth]{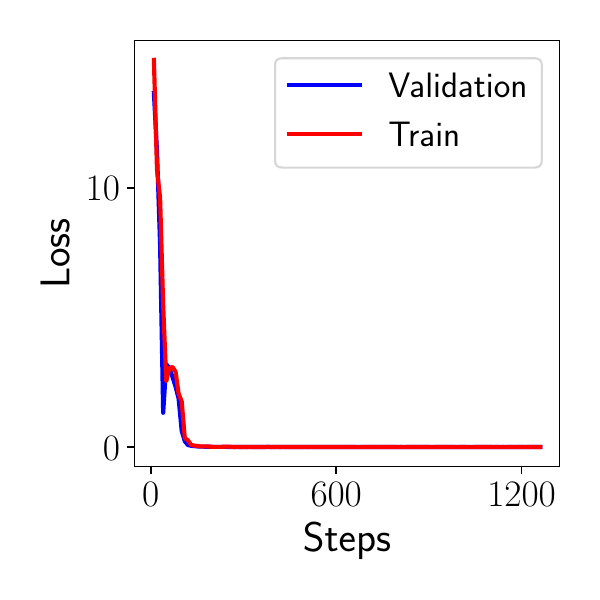}
            \captionsetup{justification=centering}
            \caption{Batch size 2 \& epoch 4}    
        \end{subfigure}
        \hfill
        \begin{subfigure}[!htbp]{0.24\textwidth}   
            \centering 
            \includegraphics[width=\textwidth]{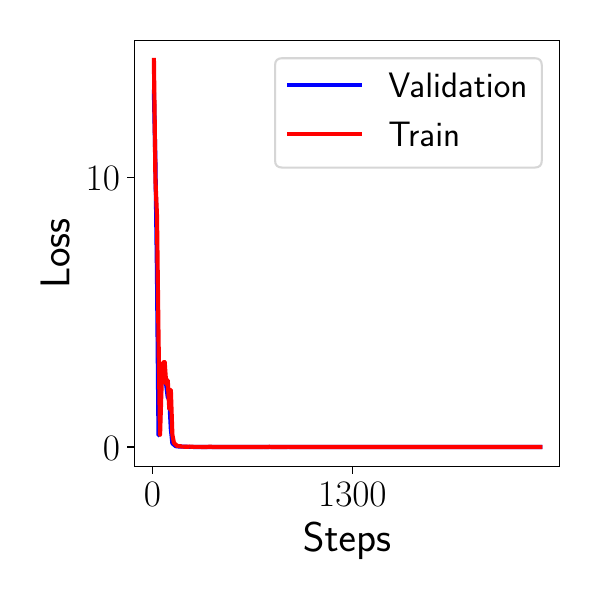}
            \captionsetup{justification=centering}
            \caption{Batch size 2 \& epoch 8}    
        \end{subfigure}
        \vskip\baselineskip
         \begin{subfigure}[!htbp]{0.24\textwidth}  
            \centering 
            \includegraphics[width=\textwidth]{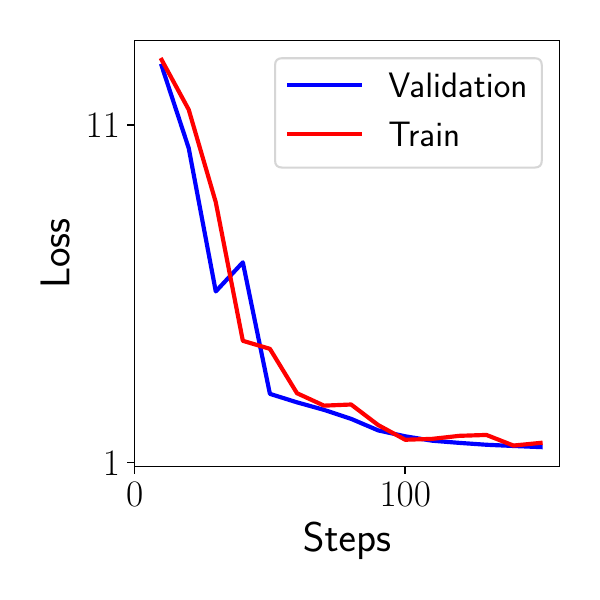}
            \captionsetup{justification=centering}
            \caption{Batch size 4 \& epoch 1}   
        \end{subfigure}
        \hfill
         \begin{subfigure}[!htbp]{0.24\textwidth}   
            \centering 
            \includegraphics[width=\textwidth]{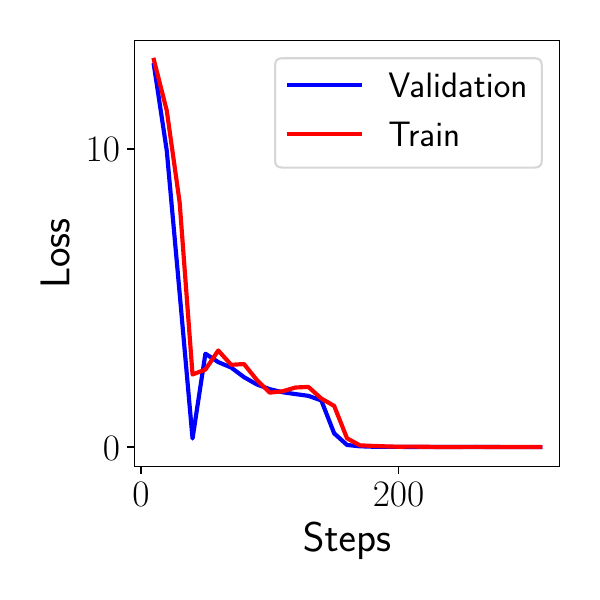}
            \captionsetup{justification=centering}
            \caption{Batch size 4 \& epoch 2}    
        \end{subfigure}
        \hfill
        \begin{subfigure}[!htbp]{0.24\textwidth}
            \centering
            \includegraphics[width=\textwidth]{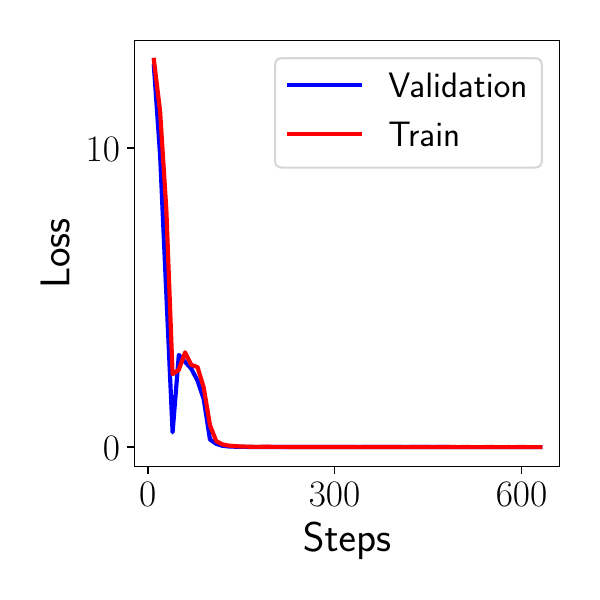}
            \captionsetup{justification=centering}
            \caption{Batch size 4 \& epoch 4}    
        \end{subfigure}
        \hfill
        \begin{subfigure}[!htbp]{0.24\textwidth}  
            \centering 
            \includegraphics[width=\textwidth]{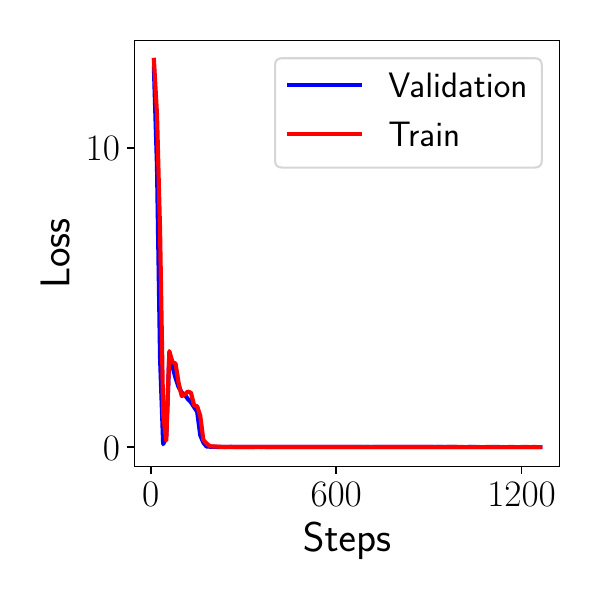}
            \captionsetup{justification=centering}
            \caption{Batch size 4 \& epoch 8}    
        \end{subfigure}
        \captionsetup{justification=centering}
        \caption{Training and validation loss convergence - Conventional Job Shop Scheduling} 
        \label{appendix:cs1}
\end{figure*}

\begin{figure*}[!htbp]
\centering
        \begin{subfigure}[!htbp]{0.24\textwidth}
            \centering
            \includegraphics[width=\textwidth]{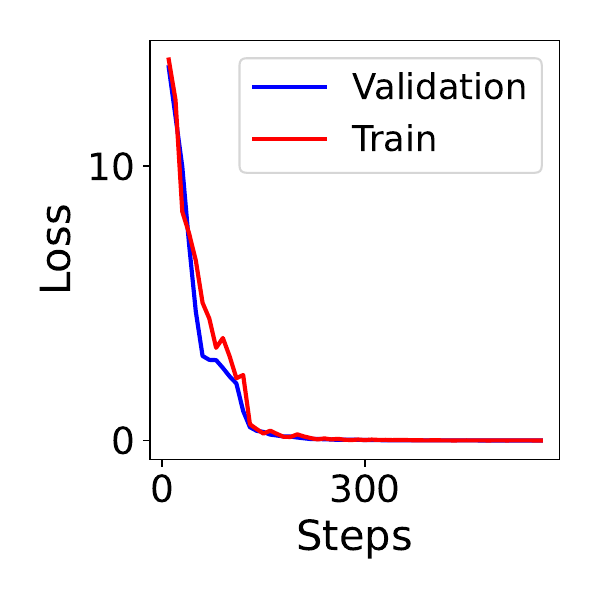}
            \captionsetup{justification=centering}
            \caption{Batch size 1 \& epoch 1}    
        \end{subfigure}
        \hfill
        \begin{subfigure}[!htbp]{0.24\textwidth}  
            \centering 
            \includegraphics[width=\textwidth]{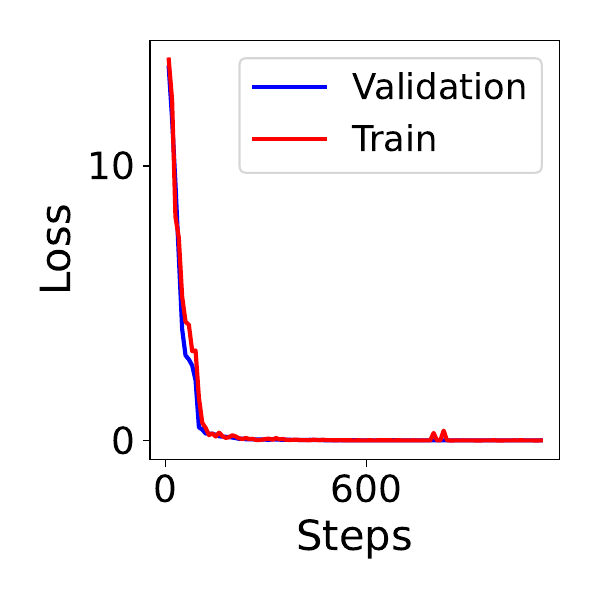}
            \captionsetup{justification=centering}
            \caption{Batch size 1 \& epoch 2}    
        \end{subfigure}
        \hfill
        \begin{subfigure}[!htbp]{0.24\textwidth}   
            \centering 
            \includegraphics[width=\textwidth]{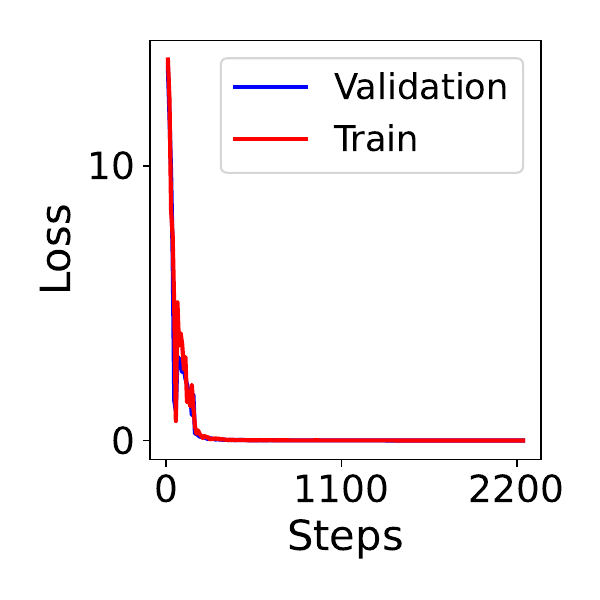}
            \captionsetup{justification=centering}
            \caption{Batch size 1 \& epoch 4}    
        \end{subfigure}
        \hfill
         \begin{subfigure}[!htbp]{0.24\textwidth}   
            \centering 
            \includegraphics[width=\textwidth]{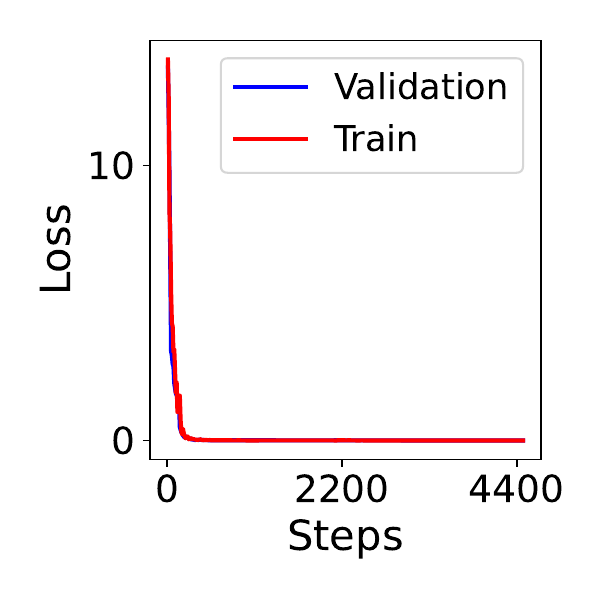}
            \captionsetup{justification=centering}
            \caption{Batch size 1 \& epoch 8}   
        \end{subfigure}
        \vskip\baselineskip
        \begin{subfigure}[!htbp]{0.24\textwidth}
            \centering
            \includegraphics[width=\textwidth]{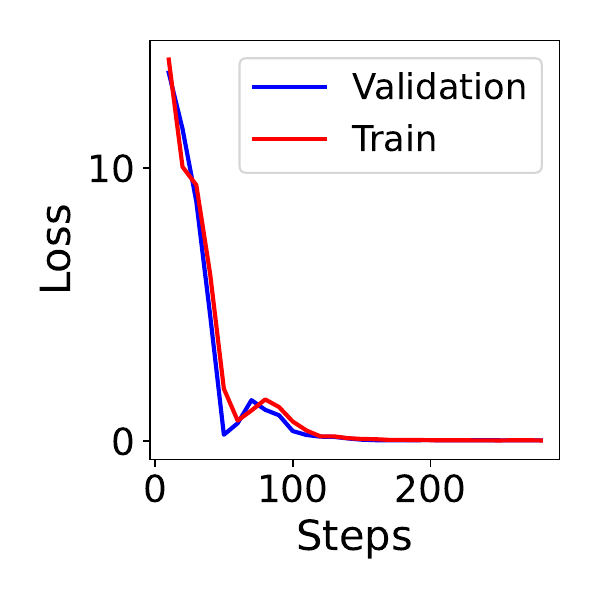}
            \captionsetup{justification=centering}
            \caption{Batch size 2 \& epoch 1}    
        \end{subfigure}
        \hfill
        \begin{subfigure}[!htbp]{0.24\textwidth}
            \centering
            \includegraphics[width=\textwidth]{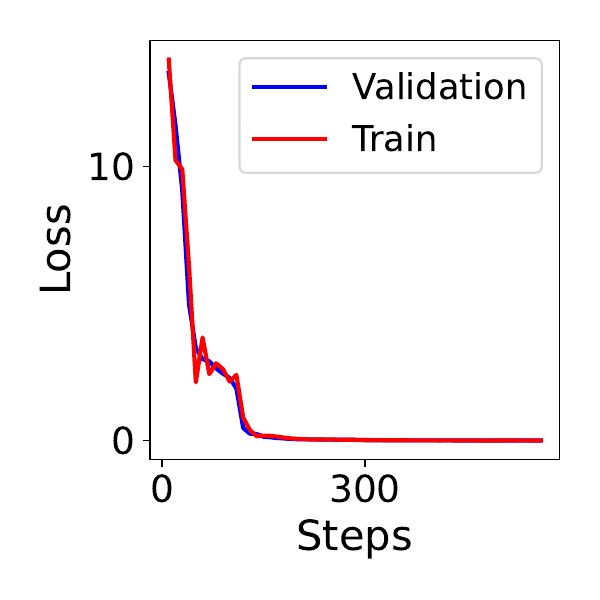}
            \captionsetup{justification=centering}
            \caption{Batch size 2 \& epoch 2}    
        \end{subfigure}
        \hfill
         \begin{subfigure}[!htbp]{0.24\textwidth}  
            \centering 
            \includegraphics[width=\textwidth]{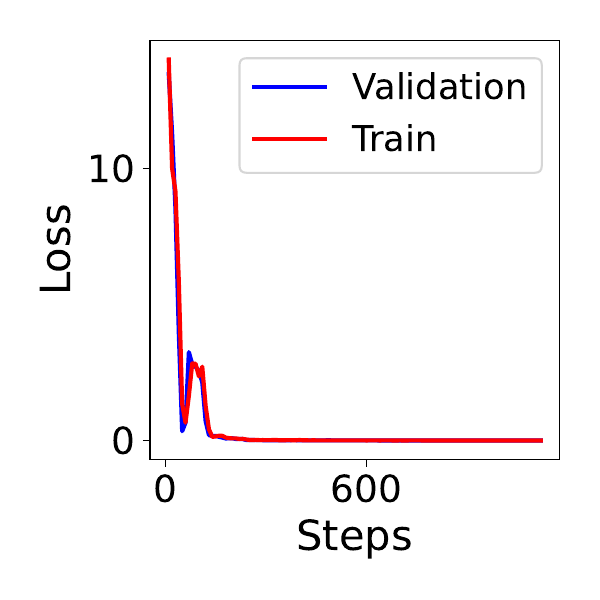}
            \captionsetup{justification=centering}
            \caption{Batch size 2 \& epoch 4}    
        \end{subfigure}
        \hfill
        \begin{subfigure}[!htbp]{0.24\textwidth}   
            \centering 
            \includegraphics[width=\textwidth]{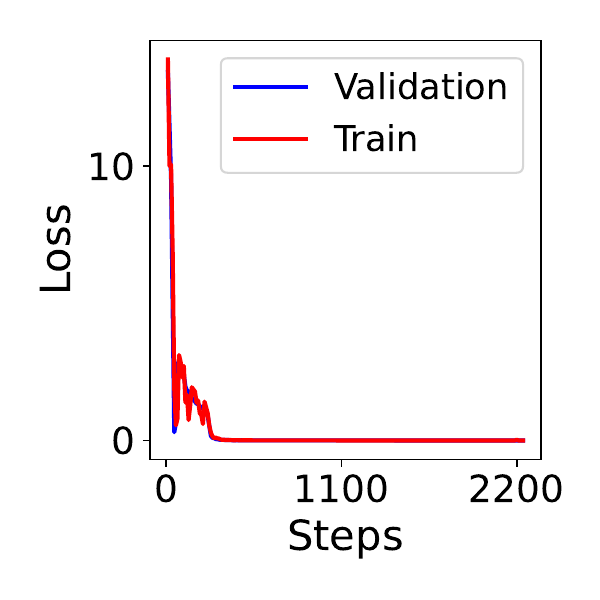}
            \captionsetup{justification=centering}
            \caption{Batch size 2 \& epoch 8}    
        \end{subfigure}
        \vskip\baselineskip
         \begin{subfigure}[!htbp]{0.24\textwidth}  
            \centering 
            \includegraphics[width=\textwidth]{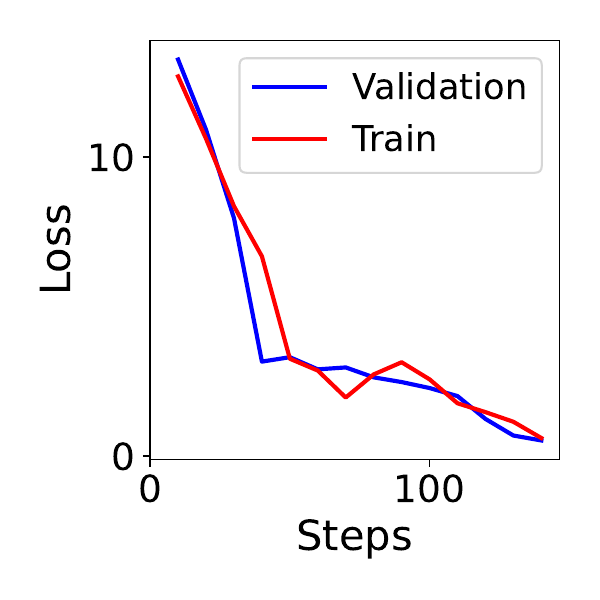}
            \captionsetup{justification=centering}
            \caption{Batch size 4 \& epoch 1}   
        \end{subfigure}
        \hfill
         \begin{subfigure}[!htbp]{0.24\textwidth}   
            \centering 
            \includegraphics[width=\textwidth]{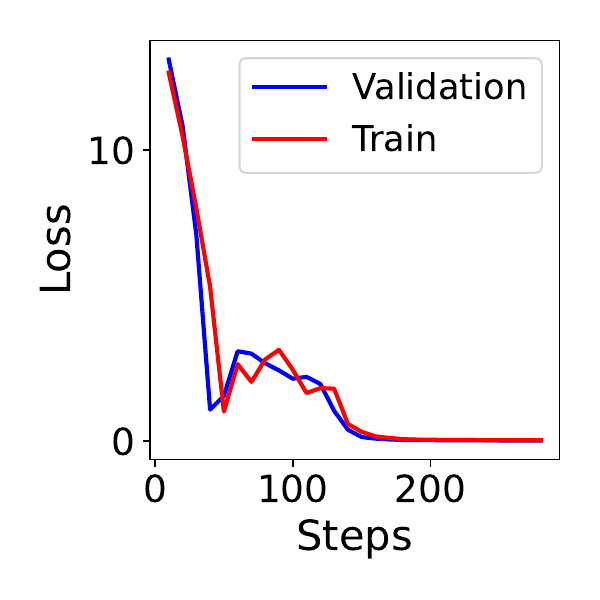}
            \captionsetup{justification=centering}
            \caption{Batch size 4 \& epoch 2}   
        \end{subfigure}
        \hfill
        \begin{subfigure}[!htbp]{0.24\textwidth}
            \centering
            \includegraphics[width=\textwidth]{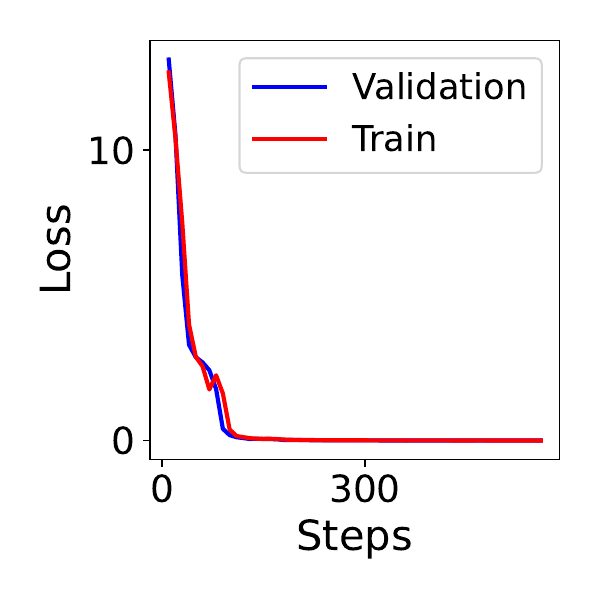}
            \captionsetup{justification=centering}
            \caption{Batch size 4 \& epoch 4}   
        \end{subfigure}
        \hfill
        \begin{subfigure}[!htbp]{0.24\textwidth}  
            \centering 
            \includegraphics[width=\textwidth]{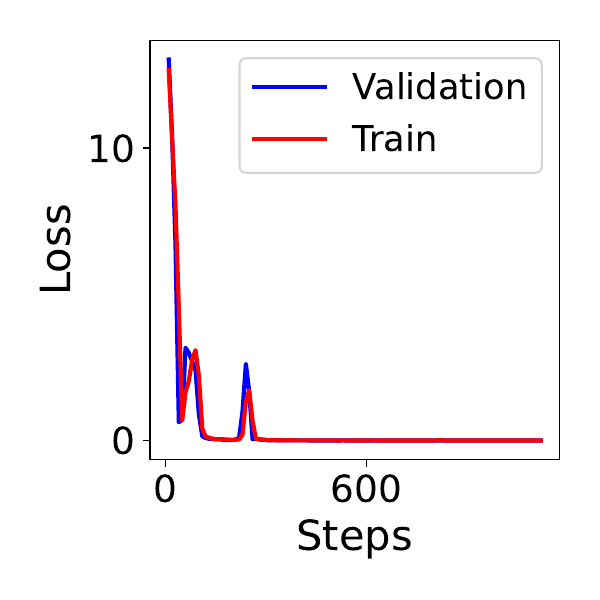}
            \captionsetup{justification=centering}
            \caption{Batch size 4 \& epoch 8}    
        \end{subfigure}
        \captionsetup{justification=centering}
        \caption{Training and validation loss convergence - Real-World Production Scheduling} 
        \label{appendix:cs2}
\end{figure*}

\bibliographystyle{unsrtnat}

\end{document}